\documentclass[11pt]{article}
\usepackage[textwidth=465pt, textheight=650pt]{geometry}

\usepackage{hyperref}       
\usepackage[utf8]{inputenc} 
\usepackage[T1]{fontenc}    
\usepackage{amsfonts}       
\usepackage{nicefrac}       
\usepackage{microtype}      
\usepackage{caption}
\usepackage{subcaption}

\usepackage[x11names]{xcolor}
\usepackage{pgfplots}
\pgfplotsset{compat=newest}
\usepackage{tikz}
\usepackage{algorithm}
\usepackage{algpseudocode}
\usepackage[acronym]{glossaries}

\newtheorem{definition}{Definition}

\usepackage{amsmath}
\usepackage{amssymb}
\usepackage{subcaption}

\usepackage[utf8]{inputenc}

\usepackage{graphicx}

\usepackage[
style=ieee,
sorting=none
]{biblatex}
\usepackage{authblk}

\title{Equality of Effort via Algorithmic Recourse}
\author[1]{
Francesca E. D. Raimondi
}
\author[1]{
Andrew R. Lawrence
}
\author[1,2]{
Hana Chockler
}

\affil[1]{causaLens}
\affil[2]{Department of Informatics, King's College London}

\date{}
\begin{document}

\maketitle

\begin{abstract}
This paper proposes a method for measuring fairness through equality of effort by applying algorithmic recourse through minimal interventions. Equality of effort is a property that can be quantified at both the individual and the group level. It answers the counterfactual question: what is the minimal cost for a protected individual or the average minimal cost for a protected group of individuals to reverse the outcome computed by an automated system?
Algorithmic recourse increases the flexibility and applicability of the notion of equal effort: it overcomes its previous limitations by reconciling multiple treatment variables, introducing feasibility and plausibility constraints, and integrating the actual relative costs of interventions. 
We extend the existing definition of equality of effort and present an algorithm for its assessment via algorithmic recourse. We validate our approach both on synthetic data and on the German credit dataset. 
\end{abstract}

\section{Introduction}
Machine Learning systems are increasingly used in many socially significant applications, such as loan approval, hiring decisions, legal processes, and healthcare, sometimes encoding existing human and historical biases, as well as generating new biases through their algorithms. Fairness is commonly defined as \textit{the absence of any prejudice or favoritism toward an individual or group based on their inherent or acquired characteristics}~\cite{Mehrabi2021:acm}.
The algorithmic fairness literature is growing quickly, but the corresponding conceptualization and applicability need further study and structure~\cite{Chouldechova2020:acm}. For instance, the prevalent correlation-based fairness algorithms fail to detect discrimination in the presence of statistical anomalies such as Simpson’s paradox~\cite{Makhlouf2020:arxiv}.  

Group fairness notions, mainly based on statistical measures, are aimed at ensuring that groups who differ in their sensitive attributes (e.g. sex, ethnicity, religion) are treated equally or at least similarly~\cite{PessacS2020:arxiv}. However group fairness, despite its suitability for policies among demographic sub-populations, does not guarantee that individuals are treated fairly and is not suitable for fine-grained groups~\cite{Makhlouf2021:elsevier}. 

In the disparate treatment liability framework (as opposed to disparate impact), discrimination claims require proof of a causal connection between the challenged decision and the sensitive feature. Therefore, \emph{causality} is becoming a fundamental tool in analyzing fairness of a decision~\cite{Makhlouf2020:arxiv}.
Causal fairness notions are not based exclusively on data but also consider the causal structure of the world, describing how data is generated and how changes in variables propagate in a system. 

Ideally, an automated system should produce the same output for every pair of individuals who differ only in their sensitive attributes. This concept is developed in~\cite{Galhotra2017:jmfse} as a measure of \textit{causal discrimination}, i.e. the fraction of inputs for which software discriminates.
More generally, \textit{fairness through awareness} requires that similar individuals will get similar outcomes~\cite{Dwork2012:itcs}. Similarity can be defined with respect to a specific task and is based on a trustworthy distance metric between individuals and a distance metric between probability distributions. This notion shows interesting developments for privacy purposes.
Similarly, \textit{individual direct discrimination} considers a target individual discriminated against if the difference observed between the rate of positive decisions for similar individuals in two groups: (1) sensitive (or protected) and (2) privileged (or unprotected), is higher than a predefined threshold~\cite{Zhang2016:ijcai}.

Using similarity measures among individuals,~\cite{Huan2020:webconf} introduces the notion of \textit{equality of effort}, that can help answer counterfactual questions like \textit{by how much should an applicant's credit score improve so that their loan application is approved?}, and judge discrimination from the perspective of equal effort among groups. Equality of effort establishes whether the effort to reach a certain outcome is the same for the protected and unprotected groups. A treatment variable is studied as to what change is required for an individual or group to achieve a certain outcome~\cite{Huan2020:webconf}.
A similar direction is pursued in~\cite{Heidari2019:arxiv} with a focus on characterizing the long-term impact of algorithmic policies on reshaping the underlying populations. 

We propose a novel way of estimating equality of effort by applying algorithmic recourse: \textit{the actions with minimal costs required for reversing unfavorable decisions by algorithms and bureaucracies across a range of counterfactual scenarios}~\cite{Karimi2021:acm}.
This has several advantages due to the flexibility of the approach in defining feasibility and plausibility constraints on variables, as well as in expressing the minimal cost required to change the outcome. In particular, algorithmic recourse, by including multiple treatment variables and integrating uniquely defined costs of interventions (e.g. relative or asymmetric), increases the expressiveness of the concept of equality of effort.
Fairness of recourse was already shown to be complementary to counterfactual fairness of prediction in \cite{VonKugelgen2022:aaai}, where the cost of recourse for an individual is compared to the cost of recourse for its counterfactual twin (i.e. a virtual individual with a counterfactual change of the sensitive variable). This opens opportunities to alternative solutions to unfairness issues by societal interventions, as opposed to the direct modification of the classifier as in \cite{Gupta2019:arxiv}.

In \S\ref{sec-EOE} and \S\ref{sec-recourse} we introduce the concepts of equality of effort and algorithmic recourse, respectively; in \S\ref{sec-eoevar} we develop our approach and outline our algorithm; in \S\ref{sec-results} we present results both on synthetic data and on the German credit dataset; finally in \S\ref{sec-conclusion} we conclude with suggested extensions to our method.

\section{Background}\label{sec-EOE}
\subsection{Equality of effort}
Unlike most causal fairness notions, which intervene on the sensitive attribute $S \notin X$, equality of effort intervenes on a treatment variable $T \in X$, where $X$ is the set of covariates. We note that any actionable variable in $X$ can be considered as a possible treatment. The definition in~\cite{Huan2020:webconf} belongs to the potential outcome framework:
\begin{definition}[$\gamma$-Minimum Effort~\cite{Huan2020:webconf}]
For individual $i$ with value $(s_i, t_i, \boldsymbol{x}_i, y_i)$, given the intervention: $Y_i(t) = (Y_i | do(t_i = t))$ and a scalar $\gamma \in [0,1]$, the minimum value of the treatment variable to achieve $\gamma$-level outcome is defined as
\footnote{$\mathbb{E}$ denotes the expectation mathematical operator.}
\begin{equation}
\Psi_i(\gamma) = \arg \min_{t \in T}\{\mathbb{E}[Y_i(t)] \geq \gamma\}
\end{equation}
and the \textbf{minimum effort} to achieve $\gamma$-level outcome is $\Psi_i(\gamma) - t_i$.
\end{definition}

The strategy proposed in~\cite{Huan2020:webconf} is based on situation testing\footnote{Situation testing is a \emph{legally grounded technique for analyzing the discriminatory treatment on individuals}~\cite{Zhang2016:ijcai}.} and the definition of a similarity to form a subset of individuals, $I = I^+ \cup I^-$, each of whom shares the same or similar characteristics ($\boldsymbol{x}$ and $t$) as individual $i$~\cite{Zhang2016:ijcai}. We denote by $I^+$ and $I^-$ the subgroups of elements in $I$ with the binary sensitive attribute value $S=s_0$ and $S=s_1$, respectively.
The derivation of the expected outcome under treatment $T=t$ of the two groups, $\mathbb{E}[Y_{I^+}(t)]$ and $\mathbb{E}[Y_{I^-}(t)]$ leads to a comparison of the $\gamma$-minimum efforts, $\Psi_{I^+}(\gamma)$ and $\Psi_{I^-}(\gamma)$, with
\begin{equation}\label{eq-EOE-gamma}
\Psi_{I^*}(\gamma) = \arg \min_{t \in T}\{\mathbb{E}[Y_{I^*}(t)] \geq \gamma\}
\end{equation}
Equality of effort is achieved when the effort discrepancy $\Psi_{I^+}(\gamma) - \Psi_{I^-}(\gamma)$ between the two sets of individuals/groups is null.
\begin{definition}[Equality of Effort]
For a given level $\gamma$, there is an equality of effort for individual $i$ if $\Psi_{I^+}(\gamma) = \Psi_{I^-}(\gamma)$.
The difference $\delta(\gamma) = \Psi_{I^+}(\gamma) - \Psi_{I^-}(\gamma)$ measures the effort discrepancy at the individual level.
\end{definition}
This definition can be extended to define effort discrepancies at a subgroup level or at the system level, if, instead of an individual $i$, we consider specific subgroups or the whole dataset, $D^+$ and $D^-$.

\begin{definition}[Average Effort Discrepancy (AED)]\label{def-AED}
If $\gamma \in \Gamma$, where $\Gamma$ is a discrete value set of the minimum expectation of the outcome variable, the Average Effort Discrepancy is defined as 
$$
AED = \frac{1}{| \Gamma |} \sum_{\gamma \in \Gamma} \delta(\gamma);  
\qquad
\text{if $\gamma \in [\gamma_1, \gamma_2]$ is continuous,}
\quad
AED = \frac{1}{\gamma_2 - \gamma_1} \int_{\gamma_1}^{\gamma_2} \delta(\gamma) d \gamma.
$$
\end{definition}

The work in~\cite{Huan2020:webconf} is based on counterfactual causal inference and develops an optimization-based framework for removing discriminatory effort unfairness from the data if discrimination is detected. 
Definition \ref{def-AED} is based on the assumptions of no hidden confounders~\cite{Pearl2009:cup}, monotonicity of the expectation of the outcome variable, invertibility of the outcome function, the presence of one binary protected attribute and one binary decision. When the invertibility assumption does not hold, causal inference methods such as outcome regression and propensity score inverse weighting are used to estimate the average treatment effect in the subgroups $I^+$ and $I^-$ .
The main limitation of equality of effort is that, generally, a single treatment variable does not appropriately reflect the unfairness between protected and unprotected groups.

\subsection{Algorithmic recourse}\label{sec-recourse}
We find the definition of recourse through minimal interventions in~\cite{Karimi2021:acm}, as opposed to recourse via nearest counterfactual explanations in~\cite{Karimi2020:pmlr}, to be particularly suited to our problem of deriving the individual cost to achieve a certain outcome.
Assuming causal sufficiency of the structural causal model $\mathcal{M}$ (i.e., no hidden confounders), additive noise, and full specification of an invertible $\mathbb{F}$ such that $\mathbb{F}(\mathbb{F}^{-1} (x) ) = x$, endogenous variable $X$ can be uniquely determined given the values of exogenous variables $U$ (and vice-versa, given the invertibility of $\mathbb{F}$).
The outcome $Y$ is related to the variables $X$ by a function $h(\cdot)$: $Y = h(X)$.

\begin{definition}[Algorithmic recourse via minimal interventions (MINT)~\cite{Karimi2021:acm}]
is an optimization problem which minimizes the cost of the set of actions $A$ (in the form of structural interventions on the factual instance $x^F$) that results in a structural counterfactual instance $x^{SCF}$ yielding the favorable output from $h$:
\begin{equation}\label{eq-MINT}
\begin{gathered}
A^*(x^F) \in \arg\min_{A} \text{cost}(A; x^F) \quad s.t. \\
h(x^F) \neq h(x^{SCF}) \\
x^{SCF} = \mathbb{F}_A ( \mathbb{F}^{-1} (x^F) ) \\
x^{SCF} \in \mathcal{P}, A \in \mathcal{F}
\end{gathered}   
\end{equation}
\end{definition}
with $cost(\cdot; x^F): \mathcal{F} \times \mathcal{X} \rightarrow \mathbb{R}_+$ and $x^{* SCF} = \mathbb{F}_{A^*} ( \mathbb{F}^{-1} (x^F) )$.
We can deterministically compute the counterfactual $x^{SCF} = \mathbb{F}_A(\mathbb{F}^{-1} (x^F) )$ by performing the Abduction-Action-Prediction steps proposed by~\cite{Pearl2009:cup}.
If $J$ is the set of variables that have been intervened upon,
\begin{equation}\label{eq-recourse-intervention}
x_j^{SCF} = [j \in J] (x_j^F + \delta_j) + [j \notin J] (x_j^F + f_j(pa_j^{SCF}) - f_j(pa_j^F))
\end{equation}
Equation \ref{eq-recourse-intervention} holds for an additive noise model with mutually independent exogenous variables.
The structural counterfactual value of the $j$-th feature, $x_j^{SCF}$, takes the value $(x_j^F + \delta_j)$ if this feature is intervened upon. Otherwise, $x_j^{SCF}$ is computed as a function of both the factual and counterfactual values of its parents, denoted respectively by $pa_j^F$ and $pa_j^{SCF}$.
The closed-form expression in Equation \ref{eq-recourse-intervention} can replace the counterfactual constraint in Equation \ref{eq-MINT}, i.e., $x^{SCF} = \mathbb{F}_A ( \mathbb{F}^{-1} (x^F) )$, after which the optimization problem in Equation \ref{eq-MINT} may be solved by building on existing frameworks for generating nearest counterfactual explanations~\cite{Karimi2020:pmlr}, including gradient-based, evolutionary-based, heuristics-based, or verification-based approaches.

An example of cost provided by~\cite{Karimi2021:acm} for Equation \ref{eq-MINT} is the $\ell_1$ norm over normalized feature changes to make effort comparable across features, i.e.
\begin{equation}\label{eq-cost}
\text{cost}(\boldsymbol{\delta}) = \sum_{j \in J} | \delta_j | / R_j, \qquad \text{where $R_j$ is the range of feature $j$.}
\end{equation}

\section{Equality of effort via algorithmic recourse}\label{sec-eoevar}
The authors in~\cite{Huan2020:webconf} limit their algorithms to changing one variable and do not consider weighted versions of the problem, representing the cost and/or the effort required in order to change the value of a variable.

The recourse-based approach has several advantages.
We can consider the treatment variable $T \in X$ in Section \ref{sec-EOE} as part of the feasibility constraints in algorithmic recourse, where we can declare which features are immutable, mutable but non-actionable, and actionable (we can intervene directly on them), and we can also add bounds to define feasible interventions (e.g. age can only increase). 
Note that in~\cite{Huan2020:webconf} only one treatment variable is considered, whereas algorithmic recourse allows us to extend the interventions to multiple treatment variables, $T \subseteq X$, included in the feasibility constraint $\mathcal{F}$.
Moreover, the calculation of the counterfactual instances allows us to propagate the effect of the change of the treatment variables throughout the structural causal model (cf. Equation \ref{eq-recourse-intervention}), and to introduce plausibility constraints on the affected covariates.
Finally, the relative costs of changing different variables can be easily reflected in the cost function as weights.

\begin{definition}[Average Minimal Effort]\label{def-average-cost}
For an individual $i$ belonging to the sensitive group $S=s_0$, we can define an Average Minimal Effort in its neighborhood subsets $I^+ \subset I$ (where $S=s_0$) and $I^- \subset I$ (where $S=s_1$):
\begin{gather}
\Phi_{I^+} = \frac{1}{|I^+|}\sum_{x \in I^+}{cost(A^*(x)) }\quad x \in I^+ \\
\Phi_{I^-}= \frac{1}{|I^-|}\sum_{x \in I^-}{cost(A^*(x))} \quad x \in I^-
\end{gather}
where $A^*$ is given by Equation \ref{eq-MINT} and $|I^*|$ denotes the cardinality of $I^*$.
\end{definition}

\begin{definition}[Equality of Effort via Algorithmic Recourse]\label{def-ACR}
We say that there is an equality of effort for an individual $i$ if $\Phi_{I^+} = \Phi_{I^-}$. The ratio $\Phi = \Phi_{I^+} / \Phi_{I^-}$ measures the \textbf{Average Cost Ratio (ACR)} at the individual level.
\end{definition}

We say that recourse is possible for an instance $x$ if there exists $A^*(x)$ that is a solution to the recourse problem in Equation \ref{eq-MINT}. 
\begin{definition}\label{def-ratio-possible-recourse}
We define the \textbf{Ratios of Possible Recourse} for an individual $i$ as
\begin{equation}\label{eq-ratio-possible-recourse}
\rho_{I^+} = \sum_{x \in I^+} \frac{\mathbb{I}_{\{ \exists A^*(x) \}}}{|I^+ |}, \qquad
\rho_{I^-}= \sum_{x \in I^-} \frac{\mathbb{I}_{\{ \exists A^*(x) \}}}{|I^- |}
\end{equation}
where $\mathbb{I}_{\{x\}}$ denotes the characteristic function of $x$.

\end{definition}

\begin{definition}\label{def-RD}
The difference $\rho = \rho_{I^-} - \rho_{I^+}$ measures the \textbf{Recourse Discrepancy (RD)} at the individual level.
\end{definition}

The RD can identify unfairness whenever a significant ratio of the elements of at least one of the two groups are so far from the decision boundaries that recourse is not a possibility for them. In fact, in this case, only considering the ACR for the instances that admit recourse could be misleading.

We can easily extend this to the subgroup level or to the system level as in~\cite{Huan2020:webconf}, if instead of individual $i$ we consider specific subgroups or the whole dataset $D^+$ and $D^-$, respectively.
For the estimation of the ACR, we propose Algorithm \ref{algo-ACR}, which has the benefit of giving an individual definition of fairness for individual $i$, if applied at an individual level.

\begin{algorithm}[!ht]
\caption{Equality of effort via algorithmic recourse}\label{algo-ACR}
\hspace*{\algorithmicindent} \textbf{Input} \\
Dataset $D = \{ D^+, D^- \}$, thresholds $\tau$, $\epsilon$, sensitive variable $S \in \{s_0, s_1\}$, unfavorable outcome $y$, feasibility constraints $\mathcal{F}$, plausibility constraints $\mathcal{P}$, individual $i$ (optional), distance $d$ (optional).  \\
\hspace*{\algorithmicindent} \textbf{Output} \\
Decision: \textit{Is there equality of effort?}
\begin{algorithmic}
\State 1. \textbf{if} individual $i$ and distance $d$ are provided as inputs \textbf{then}:
\State 2. \hspace{10 pt} Compute the similarity subsets $I = \{ I^+, I^- \}$ using distance $d$.
\State 3. \textbf{else}: 
\State 4. \hspace{10 pt} Set $I = D$, $I^+ = D^+$ and $I^- = D^-$.
\State 5. \textbf{for} each subset $I^* \in \{ I^+, I^- \}$ \textbf{do}:
\State 6. \hspace{10 pt} Identify the instances $x$ , s.t. $h(x) = y$, where $y \in \{0,1\}$ is the unfavorable outcome (i.e. the outcome we wish to change).
\State 7. \hspace{10 pt} Include and bound treatment variables $T \subseteq X$ as actionable variables in the feasibility constraint $\mathcal{F}$ and bound the covariates $X$ in the plausibility constraints $\mathcal{P}$ in Equation \ref{eq-MINT}.
\State 8. Initialize the count for the Ratio of Possible Recourse in group ${I^*}$: $\rho_{I^*} = 0$
\State 9. \hspace{10 pt} \textbf{for} $\forall x \in I^*$ \textbf{do}:
\State 10. \hspace{20 pt} Perform algorithmic recourse on $x$ following Equations \ref{eq-MINT} and \ref{eq-recourse-intervention}.
\State 11. \hspace{20 pt} \textbf{if} there exists a solution to algorithmic recourse for instance $x$ \textbf{then}:
\State 12. \hspace{30 pt} Calculate the cost of recourse for instance $x$, $\text{cost}(A^*(x))$.
\State 13. \hspace{20 pt} \textbf{else}:
\State 14. \hspace{30 pt} $\rho_{I^*} = \rho_{I^*} +\frac{1}{| I^* |}$
\State 15. \hspace{10 pt} Calculate the Average Minimal Effort as $\Phi_{I^*} = \mathbb{E}[cost(A^*(x))] \ \text{for} \ x \in I^*$ (Definition \ref{def-average-cost}).
\State 16. Compute ACR as $\Phi = \Phi_{I^+} / \Phi_{I^-}$ (Definition \ref{def-ACR}).
\State 17. Compute RD as $\rho = \rho_{I^-} - \rho_{I^+}$ (Definition \ref{def-RD}).
\State 18. \textbf{if} $| \rho | \geq \epsilon$ \textbf{then}: \textbf{return} False
\State 19. \textbf{else if} $| \Phi | \leq \tau$ \textbf{then}: \textbf{return} True
\State 20. \textbf{else}: \textbf{return} False
\end{algorithmic}
\end{algorithm}

For the sake of simplicity, we define Algorithm \ref{algo-ACR} for binary sensitive variables and for binary outcomes; however, it can be generalized to both categorical sensitive variables and outcomes. In the binary case, we only perform recourse on the instances with unfavorable outcomes.

Note that in~\cite{Huan2020:webconf}, Equation \ref{eq-EOE-gamma} relies on the estimation of the average treatment effect on outcome $Y$ ($Y$ will be favorable at least $\gamma$ fraction of the time) in the individual neighborhoods or group/dataset, and then this quantity is averaged or integrated again on all possible values of $\gamma$ in order to derive the AED in Definition \ref{def-AED}.
On the other hand, the proposed Algorithm \ref{algo-ACR} performs algorithmic recourse for each point of the cluster that has a negative outcome: our $Y$ is always flipped by the recourse for every structural counterfactual instance $x^{* SCF}$ or we find there is no feasible and plausible structural counterfactual to flip the outcome from undesired to desired as there may not always exist an action to surpass the decision boundary given the constraints. 
Our definition removes the need for further averaging or integration.

Inspired by counterfactual fairness in \cite{KusnerLRS2017:anips}, \cite{VonKugelgen2022:aaai} define the fairness of algorithmic recourse: recourse may be considered fair at the level of the individual if the cost of recourse would have been the same, had the individual belonged to a different protected group, i.e., under a counterfactual change to the sensitive variable $S$. \cite{VonKugelgen2022:aaai} show that the concepts of counterfactual fairness of prediction and counterfactual fairness of recourse are complementary.
We note that counterfactually fair algorithms do not necessarily imply fairness of recourse, and, conversely, fairness of recourse does not guarantee counterfactual fairness, as detailed in \cite{VonKugelgen2022:aaai}. Moreover, when an algorithm is counterfactually unfair (i.e. a flip of the sensitive attribute for an individual reverses an unfavorable outcome, thus removing the need for recourse), counterfactual fairness of recourse, as defined by \cite{VonKugelgen2022:aaai}, concludes maximal unfairness at the individual level. Our approach based on situation testing offers a novel way of measuring the fairness of algorithmic recourse, even when the counterfactual twin has already seen its outcome reversed.

Note that Algorithm \ref{algo-ACR} has to limit the feasibility constraints in Equation \ref{eq-MINT}: the actionable variables or treatment variables $T \subseteq X$ cannot include the sensitive variable $S$.
In the trivial and degenerate case when the treatment variable and the sensitive variable were allowed to coincide, i.e. $T=S$, the mere existence of recourse on the sensitive variable would imply counterfactual unfairness for the individual and would not carry any information about equality of effort. In fact, the notion of counterfactual fairness in~\cite{KusnerLRS2017:anips, ChocklerH2022:aaai}, albeit an individual notion, offers a different perspective in that the counterfactual change is related to a simple flip of the sensitive variable. In our case, we want to establish the cost of changing other treatment variables that is required to change the outcome.

\paragraph{The choice of the distance}
\cite{Huan2020:webconf} extend the situation testing developed in~\cite{Zhang2016:ijcai} to define the two subsets $I^+$ and $I^-$ for each assessment. However, the distance function in~\cite{Zhang2016:ijcai} is unnecessarily complex~\cite{Makhlouf2020:arxiv}.
We take the $\ell_1$ or $\ell_2$ norms if all the $K$ variables are continuous, otherwise we follow an idea detailed in~\cite{Zhang2016:arxiv}.
The distance between two domain values $x_k$ and $x_k'$ of variable $X_k, k \in \{1,\dots, K \}$ is defined as the value difference, where the normalized $\ell_1$ distance or Manhattan distance is employed for continuous/ordinal attributes, and the overlap measurement is employed for categorical attributes. 
\begin{definition}
The normalized Manhattan distance is defined as
\begin{equation}
md(x_k, x_k') = \frac{| x_k - x_k'|}{range_k} = \frac{| x_k - x_k'|}{\max_{\ell}(x_k^{\ell}) - \min_{\ell}(x_k^{\ell})}
\end{equation}
\end{definition}

\begin{definition}
The overlap measurement is defined as 
\begin{equation}
overlap(x_k, x_k') = 
\begin{cases}
0 \quad if \quad x_k = x_k' \\
1 \quad otherwise
\end{cases}
\end{equation}
\end{definition}

\begin{definition}\label{def-dist}
The difference between two points $x$, $x'$ is given by
$
d(x, x') = \sum_{k=1}^{K} dist(x_k, x_k')
$
\begin{equation}
\text{where} \qquad dist(x_k, x_k') = 
\begin{cases}
md(x_k, x_k') \quad \text{if $X_k$ is continuous/ordinal} \\
overlap(x_k, x_k') \quad  \text{if $X_k$ is categorical}
\end{cases}
\end{equation}
\end{definition}

\section{Results}\label{sec-results}
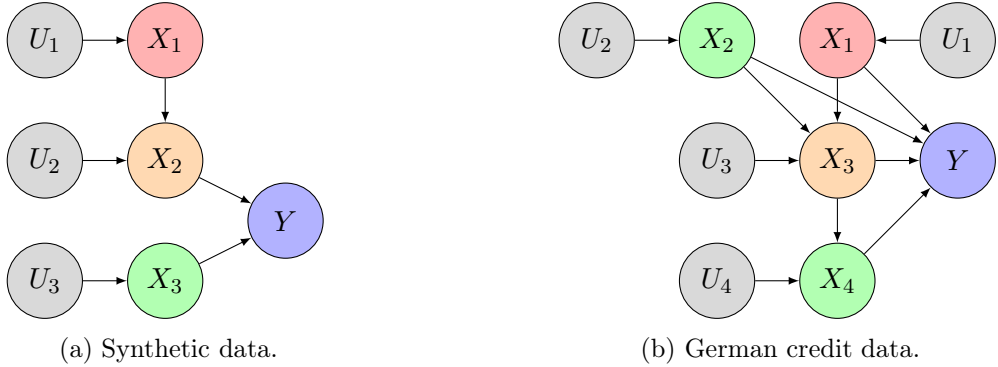
\begin{figure}[!ht]
\centering
\begin{subfigure}{0.49\textwidth}
\centering
\begin{tikzpicture}[scale=0.8,
every node/.append style={draw, circle, minimum size=1cm}]
\node[fill=gray!30] (U_1) at (0,6) {$U_1$};
\node[fill=red!30] (X_1) at (2,6) {$X_1$};
\node[fill=gray!30] (U_2) at (0,4) {$U_2$};
\node[fill=orange!30] (X_2) at (2,4) {$X_2$};
\node[fill=gray!30] (U_3) at (0,2) {$U_3$};
\node[fill=green!30] (X_3) at (2,2) {$X_3$};
\node[fill=blue!30] (Y) at (4,3) {$Y$};
\path[-latex]
(U_1) edge (X_1)
(U_2) edge (X_2)
(U_3) edge (X_3)

(X_1) edge (X_2)

(X_2) edge (Y)
(X_3) edge (Y)

;

 \end{tikzpicture}
\caption{Synthetic data.}\label{fig-DAG-synthetic}
\end{subfigure}
\begin{subfigure}{0.49\textwidth}
\centering
\begin{tikzpicture}[scale=0.8,
every node/.append style={draw, circle, minimum size=1cm}]
\node[fill=gray!30] (U_1) at (6,6) {$U_1$};
\node[fill=red!30] (X_1) at (4,6) {$X_1$};

\node[fill=gray!30] (U_2) at (0,6) {$U_2$};
\node[fill=green!30] (X_2) at (2,6) {$X_2$};

\node[fill=gray!30] (U_3) at (2,4) {$U_3$};
\node[fill=orange!30] (X_3) at (4,4) {$X_3$};

\node[fill=gray!30] (U_4) at (2,2) {$U_4$};
\node[fill=green!30] (X_4) at (4,2) {$X_4$};

\node[fill=blue!30] (Y) at (6,4) {$Y$};

\path[-latex]
(U_1) edge (X_1)
(U_2) edge (X_2)
(U_3) edge (X_3)
(U_4) edge (X_4)

(X_1) edge (X_3)

(X_2) edge (X_3)

(X_3) edge (X_4)

(X_1) edge (Y)
(X_2) edge (Y)
(X_3) edge (Y)
(X_4) edge (Y)
;

 \end{tikzpicture}
\caption{German credit data.}\label{fig-DAG-german-credit}
\end{subfigure}
\caption{Directed Acyclic Graphs (DAGs).}\label{fig-DAGS}
\end{figure}

We present results of the proposed method on synthetic data and real data from the German credit dataset~\cite{Dua2019:uci} (cf. Figure \ref{fig-DAGS}). In both cases, we learn the structural equations by fitting a linear regression model to the child-parent tuples, and a logistic regression model for the binary outcome. Distances are based on Definition \ref{def-dist} (the choice of the $\ell_1$ norm over $\ell_2$ norm for distances among neighbors does not seem to have an impact on results). For the sake of simplicity, we do not impose strict bounds on the feasibility constraints, and we fit linear relationships between variables in the causal graph. This reflects in the existence of recourse for all the individuals, meaning that we do not need to leverage the Recourse Discrepancy (RD) and we only have to consider the Average Cost Ratio (ACR) in Algorithm \ref{algo-ACR}. Also, we do not impose different relative costs on variables, but consider them equally important, i.e. we use uniform weights so the same change in each feature results in the same cost. Hence, we use Equation \ref{eq-cost} to compute costs.

\subsection{Synthetic data}
Synthetic data is composed of $n=1000$ samples following Figure \ref{fig-DAG-synthetic}. 
\begin{itemize}\setlength\itemsep{0pt}
\item The sensitive variable $X_1$ is given by the exogenous variable $U_1 \sim Bernoulli(0.5)$.
\item The proxy and actionable variable $X_2$ is given by $X_2 = \alpha \cdot X_1 + U_2$. Exogenous random variable is $U_2 \sim \mathcal{N}(3, 1)$ and parameter $\alpha$ controls the strength of the relationship with $X_1$. 
\item The actionable variable $X_3$ is given by exogenous random variable $U_3 \sim \mathcal{N}(0, 1)$.
\item The outcome is $Y = f(std(X_2 + X_3))$ where $f(\cdot)$ is the logistic sigmoid function and $std(\cdot)$ indicates standardization. 
\item $\hat{Y}$ is the predicted outcome, according to logistic regression.
\end{itemize}

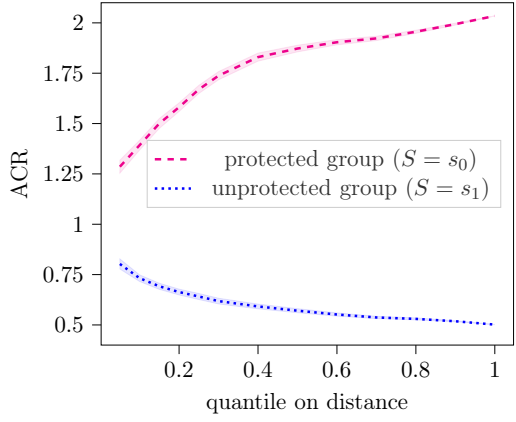
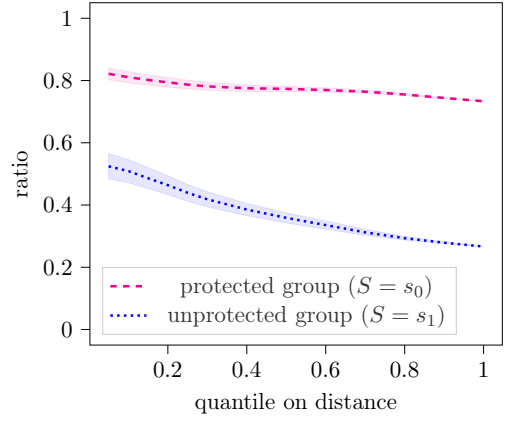
\begin{figure}[!ht]
\centering
\begin{subfigure}{0.49\textwidth}
\centering
\begin{tikzpicture}[scale=0.8]

\definecolor{darkgray176}{RGB}{176,176,176}
\definecolor{lightgray204}{RGB}{204,204,204}

\begin{axis}[
tick align=outside,
tick pos=left,
x grid style={darkgray176},
xlabel={quantile on distance},
xmin=0.0025, xmax=1.0475,
xtick style={color=black},
y grid style={darkgray176},
ylabel={ACR},
ymin=0.4, ymax=2.1,
ytick style={color=black},
ytick={0,0.25,0.5,0.75,1,1.25,1.5,1.75,2},
legend style={
  fill opacity=0.8,
  draw opacity=1,
  text opacity=1,
  at={(0.97,0.5)},
  anchor=east,
  draw=lightgray204
}
]
\path [draw=magenta, fill=magenta, opacity=0.1]
(axis cs:0.05,1.31788475644011)
--(axis cs:0.05,1.25368973083174)
--(axis cs:0.1,1.36834235736418)
--(axis cs:0.15,1.47598605876051)
--(axis cs:0.2,1.55992589298326)
--(axis cs:0.25,1.64936933969901)
--(axis cs:0.3,1.71733118960727)
--(axis cs:0.4,1.80986630642501)
--(axis cs:0.5,1.85589798787765)
--(axis cs:0.6,1.89095256087883)
--(axis cs:0.7,1.91207135699114)
--(axis cs:0.8,1.94805503118078)
--(axis cs:0.9,1.98898738088853)
--(axis cs:1,2.03444410710002)
--(axis cs:1,2.03444410710002)
--(axis cs:1,2.03444410710002)
--(axis cs:0.9,1.99838232121172)
--(axis cs:0.8,1.96371774763878)
--(axis cs:0.7,1.9336722790595)
--(axis cs:0.6,1.91668975001062)
--(axis cs:0.5,1.88896982421809)
--(axis cs:0.4,1.84994364262537)
--(axis cs:0.3,1.75794120475575)
--(axis cs:0.25,1.68962248162261)
--(axis cs:0.2,1.60259721914691)
--(axis cs:0.15,1.52502856651388)
--(axis cs:0.1,1.4171906057434)
--(axis cs:0.05,1.31788475644011)
--cycle;

\addplot [very thick, magenta, dashed]
table {%
0.0499999523162842 1.28578722476959
0.100000023841858 1.39276647567749
0.149999976158142 1.50050735473633
0.200000047683716 1.58126151561737
0.25 1.66949594020844
0.299999952316284 1.73763620853424
0.399999976158142 1.82990503311157
0.5 1.87243390083313
0.600000023841858 1.9038211107254
0.700000047683716 1.92287182807922
0.799999952316284 1.95588636398315
0.899999976158142 1.99368488788605
1 2.03444409370422
};
\addlegendentry{protected group ($S=s_0$)}

\path [draw=blue, fill=blue, opacity=0.1]
(axis cs:0.05,0.827824383277586)
--(axis cs:0.05,0.777660846353262)
--(axis cs:0.1,0.715717437344095)
--(axis cs:0.15,0.676677709295682)
--(axis cs:0.2,0.648934687929701)
--(axis cs:0.25,0.626230458980046)
--(axis cs:0.3,0.604271424928142)
--(axis cs:0.4,0.580073411967774)
--(axis cs:0.5,0.560716852571808)
--(axis cs:0.6,0.543868454865056)
--(axis cs:0.7,0.531369249271167)
--(axis cs:0.8,0.525030199275135)
--(axis cs:0.9,0.515915552415297)
--(axis cs:1,0.501642573990692)
--(axis cs:1,0.501642573990692)
--(axis cs:1,0.501642573990692)
--(axis cs:0.9,0.520768838240875)
--(axis cs:0.8,0.535028268521659)
--(axis cs:0.7,0.54291048751901)
--(axis cs:0.6,0.559815671039166)
--(axis cs:0.5,0.580764916701826)
--(axis cs:0.4,0.60368881134136)
--(axis cs:0.3,0.632612837728245)
--(axis cs:0.25,0.656170811888104)
--(axis cs:0.2,0.677783317496893)
--(axis cs:0.15,0.707048230309909)
--(axis cs:0.1,0.748623092550803)
--(axis cs:0.05,0.827824383277586)
--cycle;

\addplot [very thick, blue, dotted]
table {%
0.0499999523162842 0.802742600440979
0.100000023841858 0.732170343399048
0.149999976158142 0.691862940788269
0.200000047683716 0.663358926773071
0.25 0.641200661659241
0.299999952316284 0.618442058563232
0.399999976158142 0.591881036758423
0.5 0.570740938186646
0.600000023841858 0.551841974258423
0.700000047683716 0.537139892578125
0.799999952316284 0.530029296875
0.899999976158142 0.518342256546021
1 0.50164258480072
};
\addlegendentry{unprotected group ($S=s_1$)}
\end{axis}

\end{tikzpicture}
\caption{Average Cost Ratio (ACR).}
\label{fig-ratio-cost-synthetic}
\end{subfigure}
\hfill
\begin{subfigure}{0.49\textwidth}
\centering
\begin{tikzpicture}[scale=0.8]

\definecolor{darkgray176}{RGB}{176,176,176}
\definecolor{lightgray204}{RGB}{204,204,204}

\begin{axis}[
tick align=outside,
tick pos=left,
x grid style={darkgray176},
xlabel={quantile on distance},
xmin=0.0025, xmax=1.0475,
xtick style={color=black},
y grid style={darkgray176},
ylabel={ratio},
ymin=-0.05, ymax=1.05,
ytick style={color=black},
legend style={
  fill opacity=0.8,
  draw opacity=1,
  text opacity=1,
  at={(0.03,0.03)},
  anchor=south west,
  draw=lightgray204
},
]
\path [draw=magenta, fill=magenta, opacity=0.1]
(axis cs:0.05,0.840797809957454)
--(axis cs:0.05,0.802341712525656)
--(axis cs:0.1,0.792715610902274)
--(axis cs:0.15,0.785198463490282)
--(axis cs:0.2,0.778740045379422)
--(axis cs:0.25,0.772894467645603)
--(axis cs:0.3,0.768431028231292)
--(axis cs:0.4,0.76458643261374)
--(axis cs:0.5,0.76447731611486)
--(axis cs:0.6,0.762402120487207)
--(axis cs:0.7,0.75860960182401)
--(axis cs:0.8,0.751377311812445)
--(axis cs:0.9,0.742126645530291)
--(axis cs:1,0.733466933867734)
--(axis cs:1,0.733466933867734)
--(axis cs:1,0.733466933867734)
--(axis cs:0.9,0.745883968636524)
--(axis cs:0.8,0.758867038984826)
--(axis cs:0.7,0.769718938244229)
--(axis cs:0.6,0.776674952723669)
--(axis cs:0.5,0.782007044524865)
--(axis cs:0.4,0.785810859285212)
--(axis cs:0.3,0.793985123039215)
--(axis cs:0.25,0.800848567328122)
--(axis cs:0.2,0.809515239874373)
--(axis cs:0.15,0.818446521191845)
--(axis cs:0.1,0.828564283582184)
--(axis cs:0.05,0.840797809957454)
--cycle;

\addplot [very thick , magenta, dashed]
table {%
0.0499999523162842 0.821569681167603
0.100000023841858 0.81063985824585
0.149999976158142 0.801822423934937
0.200000047683716 0.794127702713013
0.25 0.786871433258057
0.299999952316284 0.781208038330078
0.399999976158142 0.775198698043823
0.5 0.773242235183716
0.600000023841858 0.769538521766663
0.700000047683716 0.764164209365845
0.799999952316284 0.755122184753418
0.899999976158142 0.74400532245636
1 0.733466863632202
};
\addlegendentry{protected group ($S=s_0$)}

\path [draw=blue, fill=blue, opacity=0.1]
(axis cs:0.05,0.565018347810394)
--(axis cs:0.05,0.483803815830403)
--(axis cs:0.1,0.471945177980558)
--(axis cs:0.15,0.452501056726186)
--(axis cs:0.2,0.433593947802117)
--(axis cs:0.25,0.412355773553074)
--(axis cs:0.3,0.394119349236267)
--(axis cs:0.4,0.365721909880764)
--(axis cs:0.5,0.342644137658814)
--(axis cs:0.6,0.32209110350276)
--(axis cs:0.7,0.302347429662429)
--(axis cs:0.8,0.287497795074297)
--(axis cs:0.9,0.276198240591953)
--(axis cs:1,0.266533066132265)
--(axis cs:1,0.266533066132265)
--(axis cs:1,0.266533066132265)
--(axis cs:0.9,0.281621759097284)
--(axis cs:0.8,0.29948327700729)
--(axis cs:0.7,0.321814599054443)
--(axis cs:0.6,0.349032198285926)
--(axis cs:0.5,0.374959428865556)
--(axis cs:0.4,0.405107731887915)
--(axis cs:0.3,0.4425734215428)
--(axis cs:0.25,0.466291930997513)
--(axis cs:0.2,0.493884373485813)
--(axis cs:0.15,0.519545455619568)
--(axis cs:0.1,0.545869160802876)
--(axis cs:0.05,0.565018347810394)
--cycle;

\addplot [very thick , blue, dotted]
table {%
0.0499999523162842 0.524411082267761
0.100000023841858 0.508907198905945
0.149999976158142 0.486023187637329
0.200000047683716 0.463739156723022
0.25 0.439323902130127
0.299999952316284 0.418346405029297
0.399999976158142 0.385414838790894
0.5 0.35880184173584
0.600000023841858 0.335561633110046
0.700000047683716 0.312080979347229
0.799999952316284 0.293490529060364
0.899999976158142 0.278910040855408
1 0.266533136367798
};
\addlegendentry{unprotected group ($S=s_1$)}
\end{axis}

\end{tikzpicture}
\caption{Average ratio of protected elements in $I = I^+ \cup I^-$}
\label{fig-ratio-size-synthetic}
\end{subfigure}
\caption{Equality of Effort for $\alpha=2$ - synthetic data.}
\end{figure}

Figure \ref{fig-ratio-cost-synthetic} refers to the ACR in Definition \ref{def-ACR}, whereas Figure \ref{fig-ratio-size-synthetic} refers to the ratio of protected individuals in the neighborhood $I = I^+ \cup I^-$, both as functions of the quantile threshold on distance for the subset $I$ centered on each individual. The center line refers to the average value, enclosed by the $95\%$ confidence interval. The variance tends to be higher for smaller quantiles (as there are fewer elements in each subset), whereas for unitary quantile the variance is zero as we always consider the whole dataset for comparisons and can therefore derive the effort discrepancy at system level from that point. The ACR steadily remains over $1.2$ for all radii, thus confirming the unfairness (inequality of effort) at both the individual level ($ACR(0.2) \approx 1.5$) and the system level ($ACR(1) \approx 2$). On the contrary, for the unprotected group $ACR(q) <=0.8 \ \forall q \in Q$ and $ACR(1) \approx 0.5$.
Note that the two curves of Figure \ref{fig-ratio-cost-german-credit} diverge at the system level (i.e. for high quantiles). 

\begin{figure}[!ht]
\begin{subfigure}{0.49\textwidth}
\centering
\input{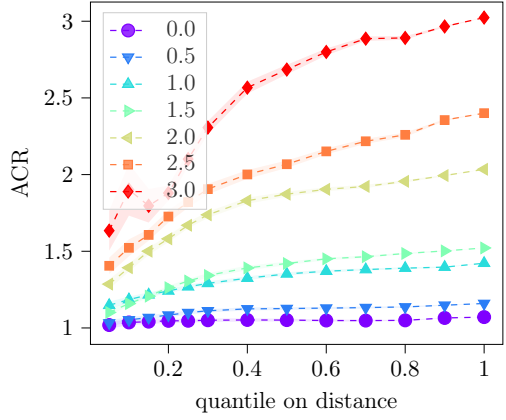}
\caption{Average Cost Ratio (ACR).}
\label{fig-ratio-cost-synthetic-alpha}
\end{subfigure}
\hfill
\begin{subfigure}{0.49\textwidth}
\centering
\input{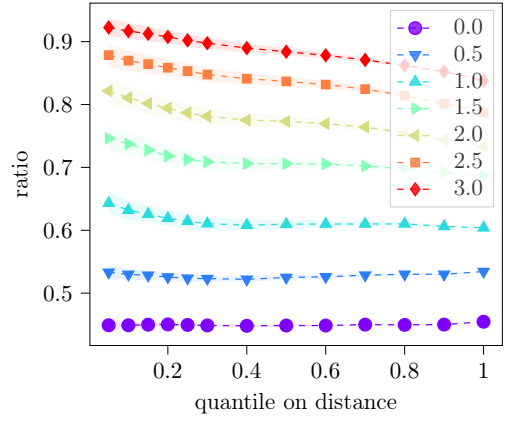}
\caption{Average ratio of protected elements in $I = I^+ \cup I^-$.}
\label{fig-ratio-size-synthetic-alpha}
\end{subfigure}
\caption{Plots for different values of $\alpha$ - synthetic data.}
\end{figure}

Figure \ref{fig-ratio-cost-synthetic-alpha} describes the ACR averaged over all the protected individuals ($S=s_0)$, whereas Figure \ref{fig-ratio-size-synthetic-alpha} presents the corresponding average ratio of elements with the sensitive attribute in the subset $I = I^+ \cup I^-$, both for different values of $\alpha$. We notice that as $\alpha$ increases, the unfairness also grows in terms of the inequality of effort. When $\alpha$ is zero, we see that ACR stays at one and we can conclude the system has an equality of effort.

\begin{figure}[!ht]
\setlength{\belowcaptionskip}{0.5\baselineskip}
\centering
\begin{subfigure}{0.49\textwidth}
\centering
\begin{tikzpicture}[scale=0.8]

\definecolor{darkgray176}{RGB}{176,176,176}
\definecolor{lightgray204}{RGB}{204,204,204}

\begin{axis}[
legend cell align={left},
legend style={
  fill opacity=0.8,
  draw opacity=1,
  text opacity=1,
  at={(0.03,0.97)},
  anchor=north west,
  draw=lightgray204
},
tick align=outside,
tick pos=left,
x grid style={darkgray176},
xlabel={quantile on distance},
xmin=0.0025, xmax=1.0475,
xtick style={color=black},
y grid style={darkgray176},
ylabel={average cost},
ymin=0.5, ymax=1.59,
ytick style={color=black}
]
\path [draw=magenta, fill=magenta, opacity=0.1]
(axis cs:0.05,1.33960080901535)
--(axis cs:0.05,1.20583637653571)
--(axis cs:0.1,1.18301715329447)
--(axis cs:0.15,1.1710769045353)
--(axis cs:0.2,1.16502033371821)
--(axis cs:0.25,1.16495180860202)
--(axis cs:0.3,1.17115484136965)
--(axis cs:0.4,1.19128406667655)
--(axis cs:0.5,1.22321158332685)
--(axis cs:0.6,1.2585235579894)
--(axis cs:0.7,1.29130459993948)
--(axis cs:0.8,1.31712729864155)
--(axis cs:0.9,1.33698519186426)
--(axis cs:1,1.35936289281524)
--(axis cs:1,1.35936289281524)
--(axis cs:1,1.35936289281524)
--(axis cs:0.9,1.34184620344837)
--(axis cs:0.8,1.32599796238328)
--(axis cs:0.7,1.30546779819252)
--(axis cs:0.6,1.27884667053212)
--(axis cs:0.5,1.2510537731342)
--(axis cs:0.4,1.22959535228565)
--(axis cs:0.3,1.22585636170412)
--(axis cs:0.25,1.23088352110885)
--(axis cs:0.2,1.24392181515395)
--(axis cs:0.15,1.26503964902162)
--(axis cs:0.1,1.29433851605223)
--(axis cs:0.05,1.33960080901535)
--cycle;

\path [draw=blue, fill=blue, opacity=0.1]
(axis cs:0.05,1.08834388540761)
--(axis cs:0.05,0.954579452927966)
--(axis cs:0.1,0.861244305708876)
--(axis cs:0.15,0.784151123574301)
--(axis cs:0.2,0.727958070775583)
--(axis cs:0.25,0.683797285741809)
--(axis cs:0.3,0.658722161231777)
--(axis cs:0.4,0.639606793010579)
--(axis cs:0.5,0.645626209411984)
--(axis cs:0.6,0.655975671859081)
--(axis cs:0.7,0.668422581854497)
--(axis cs:0.8,0.671556379673286)
--(axis cs:0.9,0.669579824747305)
--(axis cs:1,0.668174116001126)
--(axis cs:1,0.668174116001126)
--(axis cs:1,0.668174116001126)
--(axis cs:0.9,0.674440836331416)
--(axis cs:0.8,0.68042704341502)
--(axis cs:0.7,0.682585780107535)
--(axis cs:0.6,0.676298784401802)
--(axis cs:0.5,0.673468399219339)
--(axis cs:0.4,0.677918078619676)
--(axis cs:0.3,0.713423681566254)
--(axis cs:0.25,0.749728998248633)
--(axis cs:0.2,0.806859552211329)
--(axis cs:0.15,0.878113868060615)
--(axis cs:0.1,0.972565668466643)
--(axis cs:0.05,1.08834388540761)
--cycle;

\addplot [very thick , magenta, dashed]
table {%
0.0499999523162842 1.27271854877472
0.100000023841858 1.23867785930634
0.149999976158142 1.21805822849274
0.200000047683716 1.20447111129761
0.25 1.19791769981384
0.299999952316284 1.19850564002991
0.399999976158142 1.21043968200684
0.5 1.23713266849518
0.600000023841858 1.26868510246277
0.700000047683716 1.29838621616364
0.799999952316284 1.32156264781952
0.899999976158142 1.33941566944122
1 1.35936284065247
};
\addlegendentry{average cost for $I^+$}
\addplot [very thick , blue, dotted]
table {%
0.0499999523162842 1.02146172523499
0.100000023841858 0.916904926300049
0.149999976158142 0.831132411956787
0.200000047683716 0.767408847808838
0.25 0.716763138771057
0.299999952316284 0.686072945594788
0.399999976158142 0.658762454986572
0.5 0.659547328948975
0.600000023841858 0.666137218475342
0.700000047683716 0.675504207611084
0.799999952316284 0.675991773605347
1 0.668174028396606
};
\addlegendentry{average cost for $I^-$}
\end{axis}

\end{tikzpicture}
\caption{Cost of recourse, protected group ($S=s_0$).}
\label{fig-costs-synthetic-s_0}
\end{subfigure}
\hfill
\begin{subfigure}{0.49\textwidth}
\centering
\begin{tikzpicture}[scale=0.8]

\definecolor{darkgray176}{RGB}{176,176,176}
\definecolor{lightgray204}{RGB}{204,204,204}

\begin{axis}[
legend cell align={left},
legend style={
  fill opacity=0.8,
  draw opacity=1,
  text opacity=1,
  at={(0.03,0.97)},
  anchor=north west,
  draw=lightgray204
},
tick align=outside,
tick pos=left,
x grid style={darkgray176},
xlabel={quantile on distance},
xmin=0.0025, xmax=1.0475,
xtick style={color=black},
y grid style={darkgray176},
ylabel={average cost},
ymin=0.5, ymax=1.59,
ytick style={color=black},
]
\path [draw=blue, fill=blue, opacity=0.1]
(axis cs:0.05,0.700637659196685)
--(axis cs:0.05,0.571715915681059)
--(axis cs:0.1,0.577517021304802)
--(axis cs:0.15,0.587575461542808)
--(axis cs:0.2,0.601676181126035)
--(axis cs:0.25,0.614631122875579)
--(axis cs:0.3,0.624478947196823)
--(axis cs:0.4,0.641578892626844)
--(axis cs:0.5,0.656335544469753)
--(axis cs:0.6,0.668796143595466)
--(axis cs:0.7,0.676333564942113)
--(axis cs:0.8,0.68294101082251)
--(axis cs:0.9,0.680987642343176)
--(axis cs:1,0.680457458585043)
--(axis cs:1,0.680457458585043)
--(axis cs:1,0.680457458585043)
--(axis cs:0.9,0.683900133755413)
--(axis cs:0.8,0.686662539421232)
--(axis cs:0.7,0.6835841945263)
--(axis cs:0.6,0.67720899013921)
--(axis cs:0.5,0.669983251008574)
--(axis cs:0.4,0.659033015148938)
--(axis cs:0.3,0.650472021923464)
--(axis cs:0.25,0.649322155074211)
--(axis cs:0.2,0.647187055459722)
--(axis cs:0.15,0.651812237036215)
--(axis cs:0.1,0.664836583867382)
--(axis cs:0.05,0.700637659196685)
--cycle;

\path [draw=magenta, fill=magenta, opacity=0.1]
(axis cs:0.05,0.840925346980546)
--(axis cs:0.05,0.712003603464919)
--(axis cs:0.1,0.812492096653977)
--(axis cs:0.15,0.883946281318555)
--(axis cs:0.2,0.944909536634559)
--(axis cs:0.25,0.99707534473298)
--(axis cs:0.3,1.04551144335318)
--(axis cs:0.4,1.11184729408521)
--(axis cs:0.5,1.17063332782205)
--(axis cs:0.6,1.22551401348944)
--(axis cs:0.7,1.26760845246966)
--(axis cs:0.8,1.29334467602076)
--(axis cs:0.9,1.31588762101996)
--(axis cs:1,1.35645874944751)
--(axis cs:1,1.35645874944751)
--(axis cs:1,1.35645874944751)
--(axis cs:0.9,1.3188001124322)
--(axis cs:0.8,1.29706620461949)
--(axis cs:0.7,1.27485908205384)
--(axis cs:0.6,1.23392686003318)
--(axis cs:0.5,1.18428103436087)
--(axis cs:0.4,1.1293014166073)
--(axis cs:0.3,1.07150451807982)
--(axis cs:0.25,1.03176637693161)
--(axis cs:0.2,0.990420410968245)
--(axis cs:0.15,0.948183056811962)
--(axis cs:0.1,0.899811659216557)
--(axis cs:0.05,0.840925346980546)
--cycle;

\addplot [very thick , blue, dotted]
table {%
0.0499999523162842 0.636176824569702
0.100000023841858 0.621176719665527
0.149999976158142 0.619693875312805
0.200000047683716 0.624431610107422
0.25 0.63197660446167
0.299999952316284 0.637475490570068
0.5 0.663159370422363
0.600000023841858 0.673002481460571
0.700000047683716 0.679958820343018
0.799999952316284 0.684801816940308
0.899999976158142 0.682443857192993
1 0.680457472801208
};
\addlegendentry{average cost for $I^+$}
\addplot [very thick , magenta, dashed]
table {%
0.0499999523162842 0.776464462280273
0.100000023841858 0.856151819229126
0.149999976158142 0.916064739227295
0.200000047683716 0.967664957046509
0.25 1.01442086696625
0.299999952316284 1.05850803852081
0.399999976158142 1.12057435512543
0.5 1.17745721340179
0.600000023841858 1.22972047328949
0.700000047683716 1.27123379707336
0.799999952316284 1.29520547389984
0.899999976158142 1.31734383106232
1 1.35645878314972
};
\addlegendentry{average cost for $I^-$}
\end{axis}

\end{tikzpicture}
\caption{Cost of recourse, unprotected group ($S=s_1$).}
\label{fig-costs-synthetic-s_1}
\end{subfigure}
\begin{subfigure}{0.49\textwidth}
\centering
\begin{tikzpicture}[scale=0.8]

\definecolor{darkgray176}{RGB}{176,176,176}
\definecolor{lightgray204}{RGB}{204,204,204}

\begin{axis}[
legend cell align={left},
legend style={
  fill opacity=0.8,
  draw opacity=1,
  text opacity=1,
  at={(0.03,0.97)},
  anchor=north west,
  draw=lightgray204
},
tick align=outside,
tick pos=left,
x grid style={darkgray176},
xlabel={quantile on distance},
xmin=0.0025, xmax=1.0475,
xtick style={color=black},
y grid style={darkgray176},
ylabel={average subset size},
ymin=-0, ymax=430,
ytick style={color=black}
]
\path [draw=magenta, fill=magenta, opacity=0.1]
(axis cs:0.05,40.6050726478378)
--(axis cs:0.05,38.4823590461513)
--(axis cs:0.1,73.7389057146504)
--(axis cs:0.15,105.768432767979)
--(axis cs:0.2,135.00195541228)
--(axis cs:0.25,161.646675416562)
--(axis cs:0.3,186.24451465488)
--(axis cs:0.4,228.831807259229)
--(axis cs:0.5,262.344768958511)
--(axis cs:0.6,289.391464462687)
--(axis cs:0.7,312.691572814855)
--(axis cs:0.8,332.311809750797)
--(axis cs:0.9,348.862504951896)
--(axis cs:1,366)
--(axis cs:1,366)
--(axis cs:1,366)
--(axis cs:0.9,351.148424009852)
--(axis cs:0.8,336.934091888547)
--(axis cs:0.7,319.499684015746)
--(axis cs:0.6,298.264273242231)
--(axis cs:0.5,272.868345795588)
--(axis cs:0.4,239.922291101427)
--(axis cs:0.3,196.219966219437)
--(axis cs:0.25,170.560974856662)
--(axis cs:0.2,142.610066445644)
--(axis cs:0.15,111.734299472459)
--(axis cs:0.1,77.8567227006501)
--(axis cs:0.05,40.6050726478378)
--cycle;

\path [draw=blue, fill=blue, opacity=0.1]
(axis cs:0.05,9.19890949398391)
--(axis cs:0.05,7.44589924918549)
--(axis cs:0.1,15.4730631827653)
--(axis cs:0.15,23.7644701524052)
--(axis cs:0.2,32.3001611804698)
--(axis cs:0.25,41.176108901167)
--(axis cs:0.3,49.7820466439875)
--(axis cs:0.4,64.5265466001834)
--(axis cs:0.5,75.5588147365304)
--(axis cs:0.6,85.5458557324405)
--(axis cs:0.7,95.599339755858)
--(axis cs:0.8,106.995151319572)
--(axis cs:0.9,119.478471545686)
--(axis cs:1,133)
--(axis cs:1,133)
--(axis cs:1,133)
--(axis cs:0.9,121.734643208412)
--(axis cs:0.8,110.988455237805)
--(axis cs:0.7,100.805031828841)
--(axis cs:0.6,91.4486797866852)
--(axis cs:0.5,81.8892726951636)
--(axis cs:0.4,70.8341091375215)
--(axis cs:0.3,55.5786090937174)
--(axis cs:0.25,46.5233446507455)
--(axis cs:0.2,37.0659590381095)
--(axis cs:0.15,27.7929068967751)
--(axis cs:0.1,18.5433302598577)
--(axis cs:0.05,9.19890949398391)
--cycle;

\addplot [very thick , magenta, dashed]
table {%
0.0499999523162842 39.5437164306641
0.100000023841858 75.7978134155273
0.149999976158142 108.751365661621
0.200000047683716 138.806015014648
0.25 166.103820800781
0.299999952316284 191.232238769531
0.399999976158142 234.377044677734
0.5 267.606567382812
0.600000023841858 293.827880859375
0.700000047683716 316.095642089844
0.799999952316284 334.622955322266
0.899999976158142 350.005462646484
1 366
};
\addlegendentry{average subset size for $I^+$}
\addplot [very thick , blue, dotted]
table {%
0.0499999523162842 8.32240390777588
0.100000023841858 17.0081958770752
0.149999976158142 25.7786884307861
0.200000047683716 34.6830596923828
0.25 43.8497276306152
0.299999952316284 52.6803283691406
0.399999976158142 67.6803283691406
0.5 78.7240447998047
0.600000023841858 88.4972686767578
0.700000047683716 98.2021865844727
0.799999952316284 108.991806030273
0.899999976158142 120.606559753418
1 133
};
\addlegendentry{average subset size for $I^-$}
\end{axis}

\end{tikzpicture}
\caption{Size of subsets, protected group ($S=s_0$).}
\label{fig-size-subset-synthetic-s_0}
\end{subfigure}
\hfill
\begin{subfigure}{0.49\textwidth}
\centering
\begin{tikzpicture}[scale=0.8]

\definecolor{darkgray176}{RGB}{176,176,176}
\definecolor{lightgray204}{RGB}{204,204,204}

\begin{axis}[
legend cell align={left},
legend style={
  fill opacity=0.8,
  draw opacity=1,
  text opacity=1,
  at={(0.03,0.97)},
  anchor=north west,
  draw=lightgray204
},
tick align=outside,
tick pos=left,
x grid style={darkgray176},
xlabel={quantile on distance},
xmin=0.0025, xmax=1.0475,
xtick style={color=black},
y grid style={darkgray176},
ylabel={average subset size},
ymin=0, ymax=430,
ytick style={color=black},
]
\path [draw=blue, fill=blue, opacity=0.1]
(axis cs:0.05,23.9204829246146)
--(axis cs:0.05,20.3802689550847)
--(axis cs:0.1,36.6050619784656)
--(axis cs:0.15,49.0105941462255)
--(axis cs:0.2,59.0401310250766)
--(axis cs:0.25,67.1322357199639)
--(axis cs:0.3,74.6491179862339)
--(axis cs:0.4,87.3550021775449)
--(axis cs:0.5,97.4191883968929)
--(axis cs:0.6,105.680125417068)
--(axis cs:0.7,112.568137572237)
--(axis cs:0.8,119.478515346574)
--(axis cs:0.9,126.765006536092)
--(axis cs:1,133)
--(axis cs:1,133)
--(axis cs:1,133)
--(axis cs:0.9,129.054542336089)
--(axis cs:0.8,123.589153826359)
--(axis cs:0.7,118.304042878891)
--(axis cs:0.6,112.605588868646)
--(axis cs:0.5,105.016901828671)
--(axis cs:0.4,95.3818399277182)
--(axis cs:0.3,82.7042654724127)
--(axis cs:0.25,74.8076139040962)
--(axis cs:0.2,66.3583652155249)
--(axis cs:0.15,55.8164735229474)
--(axis cs:0.1,42.3122312546171)
--(axis cs:0.05,23.9204829246146)
--cycle;

\path [draw=magenta, fill=magenta, opacity=0.1]
(axis cs:0.05,23.3538312152789)
--(axis cs:0.05,19.1273717922399)
--(axis cs:0.1,37.1572153386619)
--(axis cs:0.15,55.0955619401767)
--(axis cs:0.2,72.5982984025587)
--(axis cs:0.25,90.7283984674562)
--(axis cs:0.3,108.795937606703)
--(axis cs:0.4,144.264642384465)
--(axis cs:0.5,178.831971773761)
--(axis cs:0.6,213.821546530932)
--(axis cs:0.7,251.309395004439)
--(axis cs:0.8,289.214563114342)
--(axis cs:0.9,328.554167819367)
--(axis cs:1,366)
--(axis cs:1,366)
--(axis cs:1,366)
--(axis cs:0.9,333.445832180633)
--(axis cs:0.8,299.597466960846)
--(axis cs:0.7,265.713161386538)
--(axis cs:0.6,230.704769258542)
--(axis cs:0.5,196.431186120976)
--(axis cs:0.4,162.006034307264)
--(axis cs:0.3,125.279250363221)
--(axis cs:0.25,106.008443637807)
--(axis cs:0.2,85.8678670109751)
--(axis cs:0.15,65.9119568568158)
--(axis cs:0.1,44.9781230072028)
--(axis cs:0.05,23.3538312152789)
--cycle;

\addplot [very thick , blue, dotted]
table {%
0.0499999523162842 22.1503753662109
0.100000023841858 39.4586448669434
0.149999976158142 52.4135322570801
0.200000047683716 62.6992492675781
0.25 70.9699249267578
0.299999952316284 78.6766891479492
0.399999976158142 91.3684234619141
0.5 101.218048095703
0.600000023841858 109.142860412598
0.700000047683716 115.436088562012
0.799999952316284 121.533836364746
0.899999976158142 127.909774780273
1 133
};
\addlegendentry{average subset size for $I^+$}
\addplot [very thick , magenta, dashed]
table {%
0.0499999523162842 21.2406005859375
0.100000023841858 41.0676689147949
0.149999976158142 60.5037612915039
0.200000047683716 79.2330856323242
0.25 98.3684234619141
0.299999952316284 117.03759765625
0.399999976158142 153.135345458984
0.5 187.631576538086
0.600000023841858 222.263153076172
0.700000047683716 258.511291503906
0.799999952316284 294.406005859375
0.899999976158142 331
1 366
};
\addlegendentry{average subset size for $I^-$}
\end{axis}

\end{tikzpicture}
\caption{Size of subsets, unprotected group ($S=s_1$).}
\label{fig-size-subset-synthetic-s_1}
\end{subfigure}
\caption{Details of recourse costs and subset size - synthetic data for $\alpha=2$.}
\label{fig-details-synthetic}
\end{figure}
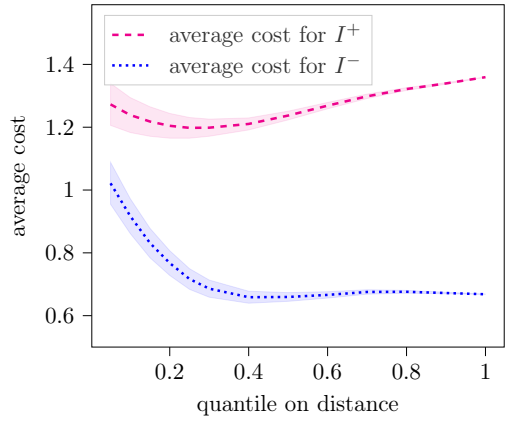
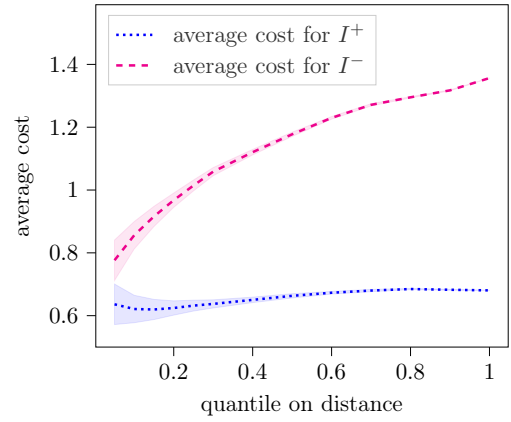
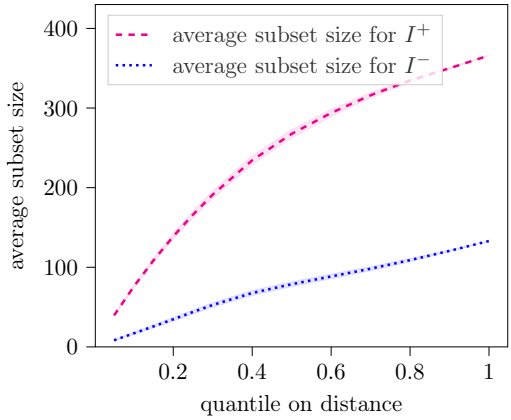
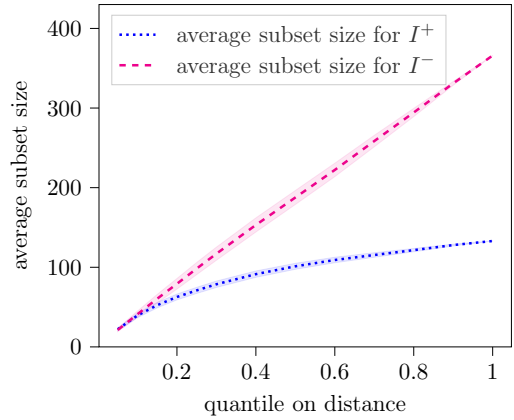

Figure \ref{fig-details-synthetic} describes in more detail the average costs for the subsets $I^+$ and $I^-$, respectively averaged over all the protected individuals ($S=s_0$) in Figure \ref{fig-costs-synthetic-s_0}) and all the unprotected individuals ($S=s_1$) in Figure \ref{fig-costs-synthetic-s_1}), both with unfavorable outcome ($Y=0$); whereas Figure \ref{fig-size-subset-synthetic-s_0} and \ref{fig-size-subset-synthetic-s_1} represent the corresponding number of elements in $I^+$ and $I^-$.

\subsection{Real data}
The German Credit dataset from the UCI repository contains $20$ attributes of $1000$ individuals applying for loans~\cite{Dua2019:uci}. The outcome is binary: a low or high credit risk, representing the likelihood of repaying the loan. We consider a subset of the features, similarly to~\cite{Karimi2021:acm}.
The setup is depicted in Figure \ref{fig-DAG-german-credit}:
\begin{itemize}\setlength\itemsep{0pt}
\item $X_1$ is the individual’s sex (the sensitive attribute, treated as immutable).
\item $X_2$ is the individual’s age (actionable but can only increase).
\item $X_3$ is the credit amount given by the bank (actionable).
\item $X_4$ is the repayment duration of the credit (non-actionable but mutable).
\item The outcome $Y$ is the ground truth for customer risk.
\item $\hat{Y}$ is the predicted customer risk, according to logistic regression.
\end{itemize}

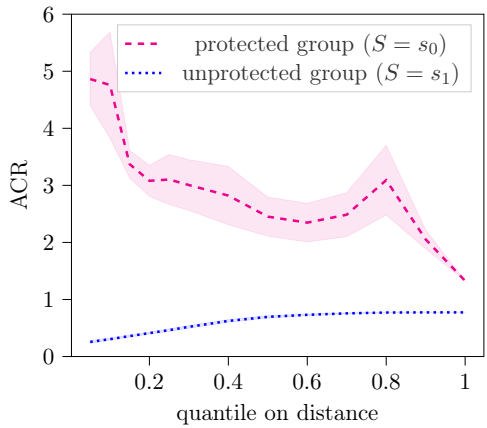
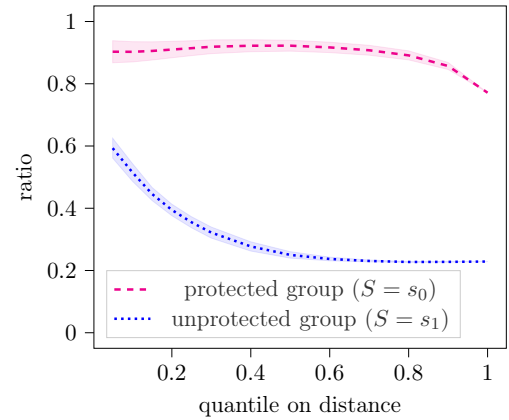
\begin{figure}[!ht]
\centering
\begin{subfigure}{0.49\textwidth}
\centering
\begin{tikzpicture}[scale=0.8]

\definecolor{darkgray176}{RGB}{176,176,176}
\definecolor{lightgray204}{RGB}{204,204,204}

\begin{axis}[
tick align=outside,
tick pos=left,
x grid style={darkgray176},
xlabel={quantile on distance},
xmin=0.0025, xmax=1.0475,
xtick style={color=black},
y grid style={darkgray176},
ylabel={ACR},
ymin=0, ymax=6,
ytick style={color=black},
ytick={0,1,2,3,4,5,6},
legend style={
  fill opacity=0.8,
  draw opacity=1,
  text opacity=1,
  at={(0.97,0.97)},
  anchor=north east,
  draw=lightgray204
}
]
\path [draw=magenta, fill=magenta, opacity=0.1]
(axis cs:0.05,5.33004421591777)
--(axis cs:0.05,4.39674199456786)
--(axis cs:0.1,3.8304133308951)
--(axis cs:0.15,3.12918000562306)
--(axis cs:0.2,2.80760165546168)
--(axis cs:0.25,2.6638084800655)
--(axis cs:0.3,2.56146408110155)
--(axis cs:0.4,2.31070492889553)
--(axis cs:0.5,2.1105999233206)
--(axis cs:0.6,2.00767049163592)
--(axis cs:0.7,2.10113085008132)
--(axis cs:0.8,2.48122956383055)
--(axis cs:0.9,1.88928115476789)
--(axis cs:1,1.32671119047703)
--(axis cs:1,1.32671119047703)
--(axis cs:1,1.32671119047703)
--(axis cs:0.9,2.2168497660311)
--(axis cs:0.8,3.70144314080272)
--(axis cs:0.7,2.87078259991184)
--(axis cs:0.6,2.6854700773554)
--(axis cs:0.5,2.79296302675361)
--(axis cs:0.4,3.33059191026601)
--(axis cs:0.3,3.44970454437142)
--(axis cs:0.25,3.54082733509258)
--(axis cs:0.2,3.35222360914259)
--(axis cs:0.15,3.61316881920745)
--(axis cs:0.1,5.68811077330438)
--(axis cs:0.05,5.33004421591777)
--cycle;

\addplot [very thick, magenta, dashed]
table {%
0.0499999523162842 4.86339330673218
0.100000023841858 4.75926208496094
0.149999976158142 3.37117433547974
0.200000047683716 3.0799126625061
0.25 3.10231781005859
0.299999952316284 3.00558423995972
0.399999976158142 2.82064843177795
0.5 2.45178151130676
0.600000023841858 2.34657025337219
0.700000047683716 2.48595666885376
0.799999952316284 3.09133625030518
0.899999976158142 2.05306553840637
1 1.32671117782593
};
\addlegendentry{protected group ($S=s_0$)}

\path [draw=blue, fill=blue, opacity=0.1]
(axis cs:0.05,0.280152882420923)
--(axis cs:0.05,0.232432798247277)
--(axis cs:0.1,0.288160851354768)
--(axis cs:0.15,0.341303961979791)
--(axis cs:0.2,0.39341163071239)
--(axis cs:0.25,0.440616537933431)
--(axis cs:0.3,0.49690323960811)
--(axis cs:0.4,0.600387903866234)
--(axis cs:0.5,0.674813223822353)
--(axis cs:0.6,0.716614176763024)
--(axis cs:0.7,0.748551123444211)
--(axis cs:0.8,0.767628046118347)
--(axis cs:0.9,0.773450928160857)
--(axis cs:1,0.773548295850518)
--(axis cs:1,0.773548295850518)
--(axis cs:1,0.773548295850518)
--(axis cs:0.9,0.773744685437833)
--(axis cs:0.8,0.774019262867725)
--(axis cs:0.7,0.764491095030919)
--(axis cs:0.6,0.745497723508048)
--(axis cs:0.5,0.715267206218252)
--(axis cs:0.4,0.647426007489201)
--(axis cs:0.3,0.543616248854337)
--(axis cs:0.25,0.485185541059151)
--(axis cs:0.2,0.427228063732885)
--(axis cs:0.15,0.375123570937527)
--(axis cs:0.1,0.320905200080917)
--(axis cs:0.05,0.280152882420923)
--cycle;

\addplot [very thick, blue, dotted]
table {%
0.0499999523162842 0.256292819976807
0.100000023841858 0.304533004760742
0.149999976158142 0.358213782310486
0.200000047683716 0.410319805145264
0.25 0.462900996208191
0.299999952316284 0.520259737968445
0.399999976158142 0.623906970024109
0.5 0.695040225982666
0.600000023841858 0.731055974960327
0.700000047683716 0.756521105766296
0.799999952316284 0.77082371711731
0.899999976158142 0.773597836494446
1 0.773548364639282
};
\addlegendentry{unprotected group ($S=s_1$)}

\end{axis}

\end{tikzpicture}
\caption{Average Cost Ratio (ACR).}
\label{fig-ratio-cost-german-credit}
\end{subfigure}
\hfill
\begin{subfigure}{0.49\textwidth}
\centering
\begin{tikzpicture}[scale=0.8]

\definecolor{darkgray176}{RGB}{176,176,176}
\definecolor{lightgray204}{RGB}{204,204,204}

\begin{axis}[
tick align=outside,
tick pos=left,
x grid style={darkgray176},
xlabel={quantile on distance},
xmin=0.0025, xmax=1.0475,
xtick style={color=black},
y grid style={darkgray176},
ylabel={ratio},
ymin=-0.05, ymax=1.05,
ytick style={color=black},
legend style={
  fill opacity=0.8,
  draw opacity=1,
  text opacity=1,
  at={(0.03,0.03)},
  anchor=south west,
  draw=lightgray204
},
]
\path [draw=magenta, fill=magenta, opacity=0.1]
(axis cs:0.05,0.93839001028789)
--(axis cs:0.05,0.867487254987772)
--(axis cs:0.1,0.870015577936447)
--(axis cs:0.15,0.876162643749476)
--(axis cs:0.2,0.883445913860607)
--(axis cs:0.25,0.890341685473279)
--(axis cs:0.3,0.89672494987592)
--(axis cs:0.4,0.902892541732269)
--(axis cs:0.5,0.904686881426696)
--(axis cs:0.6,0.900069011658339)
--(axis cs:0.7,0.891642466638594)
--(axis cs:0.8,0.876584438495423)
--(axis cs:0.9,0.845790340578828)
--(axis cs:1,0.77124183006536)
--(axis cs:1,0.77124183006536)
--(axis cs:1,0.77124183006536)
--(axis cs:0.9,0.868479385079364)
--(axis cs:0.8,0.905776253607153)
--(axis cs:0.7,0.923595444023611)
--(axis cs:0.6,0.933524164422571)
--(axis cs:0.5,0.940021697706193)
--(axis cs:0.4,0.941486322567272)
--(axis cs:0.3,0.941131006827999)
--(axis cs:0.25,0.938248841643775)
--(axis cs:0.2,0.936254100216972)
--(axis cs:0.15,0.934761177849684)
--(axis cs:0.1,0.935174861317691)
--(axis cs:0.05,0.93839001028789)
--cycle;

\addplot [very thick, magenta, dashed]
table {%
0.0499999523162842 0.902938604354858
0.100000023841858 0.902595281600952
0.149999976158142 0.90546190738678
0.25 0.914295196533203
0.299999952316284 0.918927907943726
0.399999976158142 0.922189474105835
0.5 0.922354221343994
0.600000023841858 0.916796565055847
0.700000047683716 0.907618999481201
0.799999952316284 0.891180276870728
0.899999976158142 0.857134819030762
1 0.771241903305054
};
\addlegendentry{protected group ($S=s_0$)}

\path [draw=blue, fill=blue, opacity=0.1]
(axis cs:0.05,0.625147718408287)
--(axis cs:0.05,0.561003465076544)
--(axis cs:0.1,0.486908262830331)
--(axis cs:0.15,0.424545177702546)
--(axis cs:0.2,0.376037262089696)
--(axis cs:0.25,0.337038929810178)
--(axis cs:0.3,0.303920021088287)
--(axis cs:0.4,0.262390642934188)
--(axis cs:0.5,0.239633376013495)
--(axis cs:0.6,0.231154564359142)
--(axis cs:0.7,0.22762126175922)
--(axis cs:0.8,0.226162819025259)
--(axis cs:0.9,0.22685899736336)
--(axis cs:1,0.228758169934641)
--(axis cs:1,0.228758169934641)
--(axis cs:1,0.228758169934641)
--(axis cs:0.9,0.228620067432417)
--(axis cs:0.8,0.228591399062428)
--(axis cs:0.7,0.233642494522677)
--(axis cs:0.6,0.242892810632405)
--(axis cs:0.5,0.260582923106201)
--(axis cs:0.4,0.292230659906506)
--(axis cs:0.3,0.339618148758162)
--(axis cs:0.25,0.372671628017974)
--(axis cs:0.2,0.413056385669584)
--(axis cs:0.15,0.466853792153228)
--(axis cs:0.1,0.544001017171554)
--(axis cs:0.05,0.625147718408287)
--cycle;

\addplot [very thick, blue, dotted]
table {%
0.0499999523162842 0.593075513839722
0.100000023841858 0.515454649925232
0.149999976158142 0.445699453353882
0.200000047683716 0.394546866416931
0.25 0.354855298995972
0.299999952316284 0.321769118309021
0.399999976158142 0.277310609817505
0.5 0.250108122825623
0.600000023841858 0.237023711204529
0.700000047683716 0.230631828308105
0.799999952316284 0.2273770570755
0.899999976158142 0.227739572525024
1 0.228758215904236
};
\addlegendentry{unprotected group ($S=s_1$)}

\end{axis}

\end{tikzpicture}
\caption{Average ratio of protected elements in $I = I^+ \cup I^-$.}
\label{fig-ratio-size-german-credit}
\end{subfigure}
\caption{Equality of Effort - German credit dataset.}
\end{figure}

Similarly, Figure \ref{fig-ratio-cost-german-credit} refers to the ACR in Definition \ref{def-ACR}, i.e. the ratio of the costs of the two subsets, whereas Figure \ref{fig-ratio-size-german-credit} refers to the ratio of protected individuals in the neighborhood $I = I^+ \cup I^-$.
The ACR steadily remains well over $1.2$ for all radii in the protected group, thus confirming the unfairness (inequality of effort) at both the individual level with $ACR(0.2) \approx 3$ and at the system level with $ACR(1) \approx 1.3$. 
On the contrary, for the unprotected group $ACR(q) < 0.8 \ \forall q \in Q$ and $ACR(1) \approx 0.77$.
The two curves of Figure \ref{fig-ratio-cost-german-credit} diverge at the individual level while converging at the system level, showing that group fairness notions might struggle to capture individual unfairness. 

\begin{figure}[!ht]
\setlength{\belowcaptionskip}{0.5\baselineskip}
\centering
\begin{subfigure}{0.49\textwidth}
\centering
\begin{tikzpicture}[scale=0.8]

\definecolor{darkgray176}{RGB}{176,176,176}
\definecolor{lightgray204}{RGB}{204,204,204}

\begin{axis}[
legend cell align={left},
legend style={fill opacity=0.8, draw opacity=1, text opacity=1, draw=lightgray204},
tick align=outside,
tick pos=left,
x grid style={darkgray176},
xlabel={quantile on distance},
xmin=0.0025, xmax=1.0475,
xtick style={color=black},
y grid style={darkgray176},
ylabel={average cost},
ymin=0.3, ymax=3.3,
ytick style={color=black},
ytick={0,0.5,1,1.5,2,2.5,3}
]
\path [draw=magenta, fill=magenta, opacity=0.1]
(axis cs:0.05,2.74703574635031)
--(axis cs:0.05,2.47550664947872)
--(axis cs:0.1,2.10989025575218)
--(axis cs:0.15,1.84711264960305)
--(axis cs:0.2,1.61221101839359)
--(axis cs:0.25,1.41919226233789)
--(axis cs:0.3,1.31915466779722)
--(axis cs:0.4,1.16167136276826)
--(axis cs:0.5,1.07963071301713)
--(axis cs:0.6,1.0722913304229)
--(axis cs:0.7,1.04338313849483)
--(axis cs:0.8,1.04120922864959)
--(axis cs:0.9,1.14068826051441)
--(axis cs:1,1.31489395879095)
--(axis cs:1,1.31489395879095)
--(axis cs:1,1.31489395879095)
--(axis cs:0.9,1.18361233635702)
--(axis cs:0.8,1.10761384732647)
--(axis cs:0.7,1.11334844265096)
--(axis cs:0.6,1.14875694400778)
--(axis cs:0.5,1.17320842494136)
--(axis cs:0.4,1.27838232824864)
--(axis cs:0.3,1.46628821876723)
--(axis cs:0.25,1.58508463092365)
--(axis cs:0.2,1.79711578447529)
--(axis cs:0.15,2.05213215178969)
--(axis cs:0.1,2.35390274865251)
--(axis cs:0.05,2.74703574635031)
--cycle;

\path [draw=blue, fill=blue, opacity=0.1]
(axis cs:0.05,0.775557220427901)
--(axis cs:0.05,0.504028123556311)
--(axis cs:0.1,0.499933589069327)
--(axis cs:0.15,0.555826077002714)
--(axis cs:0.2,0.552499021488032)
--(axis cs:0.25,0.545503331350605)
--(axis cs:0.3,0.549328367268907)
--(axis cs:0.4,0.554837435726945)
--(axis cs:0.5,0.554398607162569)
--(axis cs:0.6,0.568003833817373)
--(axis cs:0.7,0.550840103414677)
--(axis cs:0.8,0.540850892226704)
--(axis cs:0.9,0.660685400300609)
--(axis cs:1,0.991092837860336)
--(axis cs:1,0.991092837860336)
--(axis cs:1,0.991092837860336)
--(axis cs:0.9,0.703609476143214)
--(axis cs:0.8,0.607255510903586)
--(axis cs:0.7,0.620805407570808)
--(axis cs:0.6,0.644469447402251)
--(axis cs:0.5,0.647976319086797)
--(axis cs:0.4,0.671548401207331)
--(axis cs:0.3,0.69646191823892)
--(axis cs:0.25,0.71139569993637)
--(axis cs:0.2,0.737403787569737)
--(axis cs:0.15,0.760845579189345)
--(axis cs:0.1,0.74394608196966)
--(axis cs:0.05,0.775557220427901)
--cycle;

\addplot [very thick , magenta, dashed]
table {%
0.0499999523162842 2.61127114295959
0.100000023841858 2.23189640045166
0.149999976158142 1.94962239265442
0.200000047683716 1.70466339588165
0.25 1.50213849544525
0.299999952316284 1.39272141456604
0.399999976158142 1.22002685070038
0.5 1.12641954421997
0.600000023841858 1.11052417755127
0.700000047683716 1.07836580276489
0.799999952316284 1.0744115114212
0.899999976158142 1.16215026378632
1 1.31489396095276
};
\addlegendentry{average cost for $I^+$}
\addplot [very thick , blue, dotted]
table {%
0.0499999523162842 0.639792680740356
0.100000023841858 0.621939897537231
0.149999976158142 0.65833580493927
0.200000047683716 0.644951343536377
0.25 0.628449440002441
0.299999952316284 0.622895121574402
0.399999976158142 0.613192915916443
0.5 0.601187467575073
0.600000023841858 0.606236696243286
0.700000047683716 0.585822820663452
0.799999952316284 0.574053287506104
0.899999976158142 0.68214750289917
1 0.991092920303345
};
\addlegendentry{average cost for $I^-$}
\end{axis}

\end{tikzpicture}
\caption{Costs of recourse, protected group ($S=s_0$).}
\label{fig-costs-german-credit-s_0}
\end{subfigure}
\hfill
\begin{subfigure}{0.49\textwidth}
\centering
\begin{tikzpicture}[scale=0.8]

\definecolor{darkgray176}{RGB}{176,176,176}
\definecolor{lightgray204}{RGB}{204,204,204}

\begin{axis}[
legend cell align={left},
legend style={fill opacity=0.8, draw opacity=1, text opacity=1, draw=lightgray204},
tick align=outside,
tick pos=left,
x grid style={darkgray176},
xlabel={quantile on distance},
xmin=0.0025, xmax=1.0475,
xtick style={color=black},
y grid style={darkgray176},
ylabel={average cost},
ymin=0.3, ymax=3.3,
ytick style={color=black},
ytick={0,0.5,1,1.5,2,2.5,3}
]
\path [draw=blue, fill=blue, opacity=0.1]
(axis cs:0.05,0.873009129677756)
--(axis cs:0.05,0.712425285580891)
--(axis cs:0.1,0.815274038845563)
--(axis cs:0.15,0.870287352659259)
--(axis cs:0.2,0.896425509074948)
--(axis cs:0.25,0.917771916376466)
--(axis cs:0.3,0.93872036362308)
--(axis cs:0.4,0.963517771773459)
--(axis cs:0.5,0.978214272936713)
--(axis cs:0.6,0.983505895806424)
--(axis cs:0.7,0.995510203130441)
--(axis cs:0.8,1.00257295403202)
--(axis cs:0.9,1.00829365229319)
--(axis cs:1,1.01310591091885)
--(axis cs:1,1.01310591091885)
--(axis cs:1,1.01310591091885)
--(axis cs:0.9,1.01264510491702)
--(axis cs:0.8,1.01299974013918)
--(axis cs:0.7,1.01086217287826)
--(axis cs:0.6,1.00720670002279)
--(axis cs:0.5,1.00529136997475)
--(axis cs:0.4,0.999648438369483)
--(axis cs:0.3,0.983321601317061)
--(axis cs:0.25,0.970011098475368)
--(axis cs:0.2,0.956844571058814)
--(axis cs:0.15,0.935799085450014)
--(axis cs:0.1,0.904818875530732)
--(axis cs:0.05,0.873009129677756)
--cycle;

\path [draw=magenta, fill=magenta, opacity=0.1]
(axis cs:0.05,3.16347634980594)
--(axis cs:0.05,3.00289250570908)
--(axis cs:0.1,2.79520252413967)
--(axis cs:0.15,2.50834534534658)
--(axis cs:0.2,2.24252950695469)
--(axis cs:0.25,2.03919493245008)
--(axis cs:0.3,1.85223862793974)
--(axis cs:0.4,1.57744628486292)
--(axis cs:0.5,1.42463774002012)
--(axis cs:0.6,1.35381043643582)
--(axis cs:0.7,1.31937834921803)
--(axis cs:0.8,1.30220558824296)
--(axis cs:0.9,1.3040204453831)
--(axis cs:1,1.30968669487525)
--(axis cs:1,1.30968669487525)
--(axis cs:1,1.30968669487525)
--(axis cs:0.9,1.30837189800693)
--(axis cs:0.8,1.31263237435013)
--(axis cs:0.7,1.33473031896585)
--(axis cs:0.6,1.37751124065218)
--(axis cs:0.5,1.45171483705816)
--(axis cs:0.4,1.61357695145894)
--(axis cs:0.3,1.89683986563372)
--(axis cs:0.25,2.09143411454898)
--(axis cs:0.2,2.30294856893856)
--(axis cs:0.15,2.57385707813733)
--(axis cs:0.1,2.88474736082484)
--(axis cs:0.05,3.16347634980594)
--cycle;

\addplot [very thick , blue, dotted]
table {%
0.0499999523162842 0.792717218399048
0.100000023841858 0.86004638671875
0.149999976158142 0.903043270111084
0.200000047683716 0.926635026931763
0.299999952316284 0.961020946502686
0.399999976158142 0.981583118438721
0.5 0.991752862930298
0.600000023841858 0.995356321334839
0.700000047683716 1.00318622589111
0.799999952316284 1.00778639316559
1 1.01310586929321
};
\addlegendentry{average cost for $I^+$}
\addplot [very thick , magenta, dashed]
table {%
0.0499999523162842 3.08318448066711
0.100000023841858 2.83997488021851
0.149999976158142 2.5411012172699
0.200000047683716 2.27273893356323
0.25 2.06531453132629
0.299999952316284 1.87453925609589
0.399999976158142 1.59551167488098
0.5 1.43817627429962
0.600000023841858 1.36566078662872
0.700000047683716 1.32705438137054
0.799999952316284 1.30741894245148
0.899999976158142 1.30619621276855
1 1.3096866607666
};
\addlegendentry{average cost for $I^-$}
\end{axis}

\end{tikzpicture}
\caption{Costs of recourse, unprotected group ($S=s_1$).}
\label{fig-costs-german-credit-s_1}
\end{subfigure}
\begin{subfigure}{0.49\textwidth}
\centering
\begin{tikzpicture}[scale=0.8]

\definecolor{darkgray176}{RGB}{176,176,176}
\definecolor{lightgray204}{RGB}{204,204,204}

\begin{axis}[
legend cell align={left},
legend style={
  fill opacity=0.8,
  draw opacity=1,
  text opacity=1,
  at={(0.03,0.97)},
  anchor=north west,
  draw=lightgray204
},
tick align=outside,
tick pos=left,
x grid style={darkgray176},
xlabel={quantile on distance},
xmin=0.0025, xmax=1.0475,
xtick style={color=black},
y grid style={darkgray176},
ylabel={average subset size},
ymin=-5, ymax=155,
ytick style={color=black},
ytick={0,20,40,60,80,100,120,140}
]
\path [draw=magenta, fill=magenta, opacity=0.1]
(axis cs:0.05,15.5576603778364)
--(axis cs:0.05,13.6457294526721)
--(axis cs:0.1,24.7983835452592)
--(axis cs:0.15,34.7874211180948)
--(axis cs:0.2,43.6633332860235)
--(axis cs:0.25,51.7750546235716)
--(axis cs:0.3,59.3012369407153)
--(axis cs:0.4,71.2658860867458)
--(axis cs:0.5,80.0356508203311)
--(axis cs:0.6,87.3410671033438)
--(axis cs:0.7,94.4713245637711)
--(axis cs:0.8,101.339770979259)
--(axis cs:0.9,108.939979692959)
--(axis cs:1,118)
--(axis cs:1,118)
--(axis cs:1,118)
--(axis cs:0.9,111.076969459584)
--(axis cs:0.8,105.185652749555)
--(axis cs:0.7,99.1557940802967)
--(axis cs:0.6,92.6589328966562)
--(axis cs:0.5,85.9812983322112)
--(axis cs:0.4,77.225639336983)
--(axis cs:0.3,64.9191020423355)
--(axis cs:0.25,56.9707080882928)
--(axis cs:0.2,48.3705650190613)
--(axis cs:0.15,38.9074941361425)
--(axis cs:0.1,27.9812774716899)
--(axis cs:0.05,15.5576603778364)
--cycle;

\path [draw=blue, fill=blue, opacity=0.1]
(axis cs:0.05,3.6738190247327)
--(axis cs:0.05,1.5126216532334)
--(axis cs:0.1,2.6383120744284)
--(axis cs:0.15,3.53780611238124)
--(axis cs:0.2,4.24730798946015)
--(axis cs:0.25,4.81212417620283)
--(axis cs:0.3,5.24723323199136)
--(axis cs:0.4,6.21887744221345)
--(axis cs:0.5,7.14258646518396)
--(axis cs:0.6,8.32180853999401)
--(axis cs:0.7,9.88713941142181)
--(axis cs:0.8,12.3767286526287)
--(axis cs:0.9,17.596647224218)
--(axis cs:1,35)
--(axis cs:1,35)
--(axis cs:1,35)
--(axis cs:0.9,21.403352775782)
--(axis cs:0.8,16.8944577880492)
--(axis cs:0.7,14.5874368597646)
--(axis cs:0.6,12.9832762057687)
--(axis cs:0.5,11.7048711619347)
--(axis cs:0.4,10.5777327272781)
--(axis cs:0.3,9.43073286970356)
--(axis cs:0.25,8.81499446786496)
--(axis cs:0.2,8.05777675630256)
--(axis cs:0.15,7.03846507405944)
--(axis cs:0.1,5.71762012896143)
--(axis cs:0.05,3.6738190247327)
--cycle;

\addplot [very thick, magenta, dashed]
table {%
0.0499999523162842 14.60169506073
0.100000023841858 26.3898296356201
0.149999976158142 36.8474578857422
0.200000047683716 46.0169486999512
0.25 54.3728828430176
0.299999952316284 62.1101684570312
0.399999976158142 74.2457656860352
0.5 83.0084762573242
0.600000023841858 90
0.700000047683716 96.8135604858398
0.799999952316284 103.262710571289
0.899999976158142 110.008476257324
1 118
};
\addlegendentry{average subset size for $I^+$}
\addplot [very thick, blue, dotted]
table {%
0.0499999523162842 2.59322023391724
0.100000023841858 4.17796611785889
0.149999976158142 5.28813552856445
0.200000047683716 6.15254259109497
0.25 6.81355953216553
0.299999952316284 7.33898305892944
0.399999976158142 8.39830493927002
0.5 9.42372894287109
0.600000023841858 10.6525421142578
0.700000047683716 12.2372884750366
0.799999952316284 14.6355934143066
0.899999976158142 19.5
1 35
};
\addlegendentry{average subset size for $I^-$}
\end{axis}

\end{tikzpicture}
\caption{Size of subsets, protected group ($S=s_0$).}
\label{fig-size-subset-german-credit-s_0}
\end{subfigure}
\hfill
\begin{subfigure}{0.49\textwidth}
\centering
\begin{tikzpicture}[scale=0.8]

\definecolor{darkgray176}{RGB}{176,176,176}
\definecolor{lightgray204}{RGB}{204,204,204}

\begin{axis}[
legend cell align={left},
legend style={
  fill opacity=0.8,
  draw opacity=1,
  text opacity=1,
  at={(0.03,0.97)},
  anchor=north west,
  draw=lightgray204
},
tick align=outside,
tick pos=left,
x grid style={darkgray176},
xlabel={quantile on distance},
xmin=0.0025, xmax=1.0475,
xtick style={color=black},
y grid style={darkgray176},
ylabel={average subset size},
ymin=-5, ymax=155,
ytick style={color=black},
ytick={0,20,40,60,80,100,120,140}
]
\path [draw=blue, fill=blue, opacity=0.1]
(axis cs:0.05,19.4453723008996)
--(axis cs:0.05,16.4974848419575)
--(axis cs:0.1,23.1612996842398)
--(axis cs:0.15,26.6464786110638)
--(axis cs:0.2,28.6539840677554)
--(axis cs:0.25,29.9258029444831)
--(axis cs:0.3,30.7579238830794)
--(axis cs:0.4,32.1699757729212)
--(axis cs:0.5,32.971401559022)
--(axis cs:0.6,33.4550250950192)
--(axis cs:0.7,33.9039659855139)
--(axis cs:0.8,34.2204622378681)
--(axis cs:0.9,34.5251211991628)
--(axis cs:1,35)
--(axis cs:1,35)
--(axis cs:1,35)
--(axis cs:0.9,34.9605930865515)
--(axis cs:0.8,34.8081091907033)
--(axis cs:0.7,34.6674625859147)
--(axis cs:0.6,34.5449749049808)
--(axis cs:0.5,34.228598440978)
--(axis cs:0.4,33.7728813699359)
--(axis cs:0.3,32.9563618312063)
--(axis cs:0.25,32.3027684840883)
--(axis cs:0.2,31.4603016465303)
--(axis cs:0.15,29.7535213889362)
--(axis cs:0.1,26.7815574586174)
--(axis cs:0.05,19.4453723008996)
--cycle;

\path [draw=magenta, fill=magenta, opacity=0.1]
(axis cs:0.05,13.0274863325681)
--(axis cs:0.05,11.1439422388604)
--(axis cs:0.1,21.8886265657901)
--(axis cs:0.15,33.4052751274268)
--(axis cs:0.2,44.1627503064945)
--(axis cs:0.25,54.138216434509)
--(axis cs:0.3,64.4829261915194)
--(axis cs:0.4,83.0945276585979)
--(axis cs:0.5,97.915346940905)
--(axis cs:0.6,107.275670409101)
--(axis cs:0.7,113.12867376919)
--(axis cs:0.8,116.963633462564)
--(axis cs:0.9,117.60648635859)
--(axis cs:1,118)
--(axis cs:1,118)
--(axis cs:1,118)
--(axis cs:0.9,117.99351364141)
--(axis cs:0.8,117.550652251721)
--(axis cs:0.7,115.785611945096)
--(axis cs:0.6,112.438615305185)
--(axis cs:0.5,106.198938773381)
--(axis cs:0.4,92.6197580556879)
--(axis cs:0.3,72.8885023799091)
--(axis cs:0.25,60.9474978512053)
--(axis cs:0.2,48.7515354077912)
--(axis cs:0.15,36.7661534440018)
--(axis cs:0.1,24.5685162913528)
--(axis cs:0.05,13.0274863325681)
--cycle;

\addplot [very thick, blue, dotted]
table {%
0.0499999523162842 17.9714279174805
0.100000023841858 24.9714279174805
0.149999976158142 28.2000007629395
0.200000047683716 30.0571422576904
0.25 31.1142864227295
0.299999952316284 31.8571434020996
0.399999976158142 32.9714279174805
0.5 33.5999984741211
0.600000023841858 34
0.700000047683716 34.2857131958008
0.899999976158142 34.7428588867188
1 35
};
\addlegendentry{average subset size for $I^+$}
\addplot [very thick, magenta, dashed]
table {%
0.0499999523162842 12.08571434021
0.100000023841858 23.2285709381104
0.149999976158142 35.0857124328613
0.200000047683716 46.4571418762207
0.299999952316284 68.6857147216797
0.399999976158142 87.8571395874023
0.5 102.057144165039
0.600000023841858 109.857139587402
0.700000047683716 114.457145690918
0.799999952316284 117.257141113281
0.899999976158142 117.800003051758
1 118
};
\addlegendentry{average subset size for $I^-$}
\end{axis}

\end{tikzpicture}
\caption{Size of subsets, unprotected group ($S=s_1$).}
\label{fig-size-subset-german-credit-s_1}
\end{subfigure}
\caption{Details of recourse costs and subset size - German credit dataset.}
\label{fig-details-german-credit}
\end{figure}
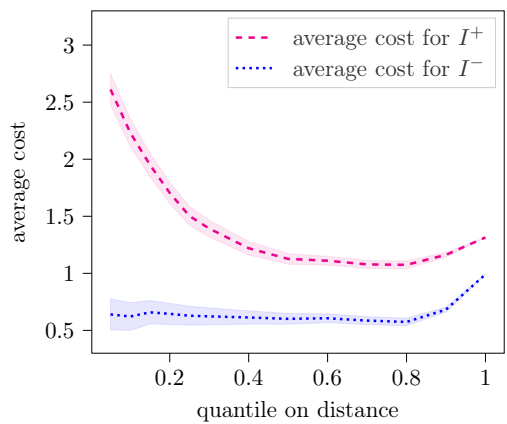
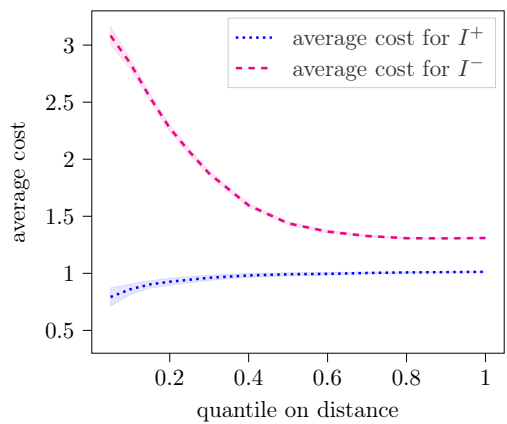
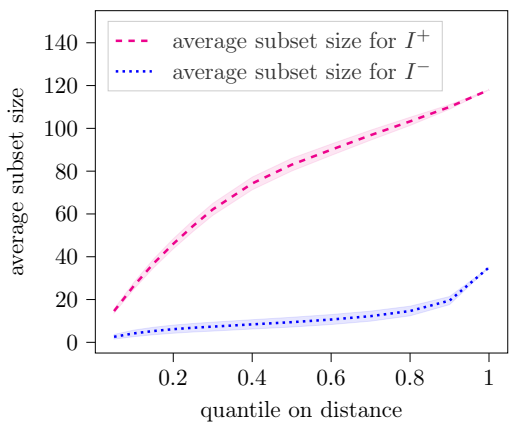
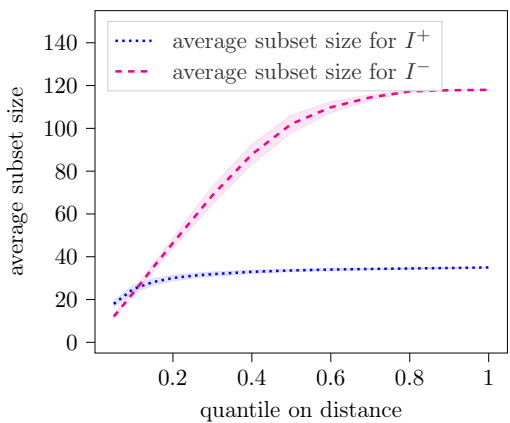

Figure \ref{fig-details-german-credit} describes in more detail the average costs for the subsets $I^+$ and $I^-$, respectively averaged over all the protected individuals ($S=s_0$) in Figure \ref{fig-costs-german-credit-s_0}) and all the unprotected individuals ($S=s_1$) in Figure \ref{fig-costs-german-credit-s_1}), both with unfavorable outcome ($Y=0$); whereas Figure \ref{fig-size-subset-german-credit-s_0} and \ref{fig-size-subset-german-credit-s_1} represents the corresponding number of elements in $I^+$ and $I^-$.

\clearpage
\subsection{Comparison between counterfactual fairness and equality of effort}
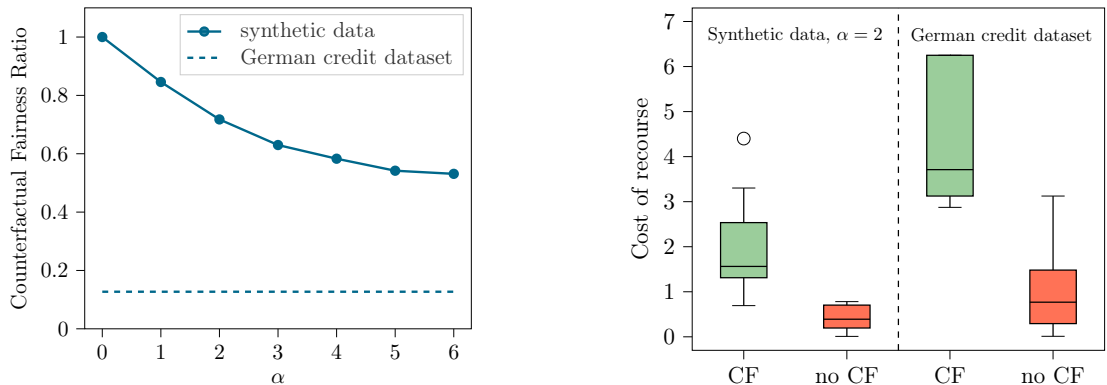
\begin{figure}[!ht]
\setlength{\belowcaptionskip}{-0.4\baselineskip}
\centering
\begin{subfigure}{0.49\textwidth}
\centering
\begin{tikzpicture}[scale=0.75]

\definecolor{darkgray176}{RGB}{176,176,176}
\definecolor{lightgray204}{RGB}{204,204,204}

\begin{axis}[
legend cell align={left},
legend style={fill opacity=0.8, draw opacity=1, text opacity=1, draw=lightgray204},
tick align=outside,
tick pos=left,
x grid style={darkgray176},
xlabel={$\alpha$},
xmin=-0.3, xmax=6.3,
xtick style={color=black},
y grid style={darkgray176},
ylabel={Counterfactual Fairness Ratio},
ymin=0, ymax=1.1,
ytick style={color=black}
]
\addplot [very thick, DeepSkyBlue4, mark=*, mark size=2, mark options={%
fill=DeepSkyBlue4,draw=DeepSkyBlue4
}]
table {%
0 1
1 0.845921516418457
2 0.717616558074951
3 0.629885077476501
4 0.58297872543335
5 0.541832685470581
6 0.531128406524658
};
\addlegendentry{synthetic data}
\addplot [very thick, DeepSkyBlue4, dashed]
table {%
0 0.127118587493896
6 0.127118587493896
};
\addlegendentry{German credit dataset}
\end{axis}

\end{tikzpicture}
\caption{CFR = 1 for complete fairness.}
\label{fig-CF-score}
\end{subfigure}
\hfill
\begin{subfigure}{0.49\textwidth}
\centering
\begin{tikzpicture}[scale=0.8]

\definecolor{darkgray176}{RGB}{176,176,176}

\begin{axis}[
tick align=outside,
tick pos=left,
x grid style={darkgray176},
xmin=0.5, xmax=4.5,
xtick style={color=black},
xtick={1,2,3,4},
xticklabels={CF ,no CF,CF,no CF},
y grid style={darkgray176},
ylabel={Cost of recourse},
ymin=-0.3, ymax=7.3,
ytick style={color=black},
ytick={0,1,2,3,4,5,6,7},
]
\path [draw=black, fill=DarkSeaGreen3, semithick]
(axis cs:0.775,1.31164003889128)
--(axis cs:1.225,1.31164003889128)
--(axis cs:1.225,2.53533771960422)
--(axis cs:0.775,2.53533771960422)
--(axis cs:0.775,1.31164003889128)
--cycle;
\addplot [semithick, black]
table {%
1 1.31164003889128
1 0.692462531858233
};
\addplot [semithick, black]
table {%
1 2.53533771960422
1 3.30309952529435
};
\addplot [black]
table {%
0.8875 0.692462531858233
1.1125 0.692462531858233
};
\addplot [black]
table {%
0.8875 3.30309952529435
1.1125 3.30309952529435
};
\addplot [black, mark=o, mark size=3, mark options={solid,fill opacity=0}, only marks]
table {%
1 4.40107739024523
};
\path [draw=black, fill=Coral1, semithick]
(axis cs:1.775,0.1953125)
--(axis cs:2.225,0.1953125)
--(axis cs:2.225,0.703744172640949)
--(axis cs:1.775,0.703744172640949)
--(axis cs:1.775,0.1953125)
--cycle;
\addplot [semithick, black]
table {%
2 0.1953125
2 0.0094448822362894
};
\addplot [semithick, black]
table {%
2 0.703744172640949
2 0.78125
};
\addplot [black]
table {%
1.8875 0.0094448822362894
2.1125 0.0094448822362894
};
\addplot [black]
table {%
1.8875 0.78125
2.1125 0.78125
};
\path [draw=black, fill=DarkSeaGreen3, semithick]
(axis cs:2.775,3.125)
--(axis cs:3.225,3.125)
--(axis cs:3.225,6.25)
--(axis cs:2.775,6.25)
--(axis cs:2.775,3.125)
--cycle;
\addplot [semithick, black]
table {%
3 3.125
3 2.87377187068025
};
\addplot [semithick, black]
table {%
3 6.25
3 6.25
};
\addplot [black]
table {%
2.8875 2.87377187068025
3.1125 2.87377187068025
};
\addplot [black]
table {%
2.8875 6.25
3.1125 6.25
};
\path [draw=black, fill=Coral1, semithick]
(axis cs:3.775,0.293250761416691)
--(axis cs:4.225,0.293250761416691)
--(axis cs:4.225,1.48115475909849)
--(axis cs:3.775,1.48115475909849)
--(axis cs:3.775,0.293250761416691)
--cycle;
\addplot [semithick, black]
table {%
4 0.293250761416691
4 0.0113105625621074
};
\addplot [semithick, black]
table {%
4 1.48115475909849
4 3.125
};
\addplot [black]
table {%
3.8875 0.0113105625621074
4.1125 0.0113105625621074
};
\addplot [black]
table {%
3.8875 3.125
4.1125 3.125
};
\addplot [semithick, black, dashed]
table {%
2.5 -0.302582873651896
2.5 7.3
};
\addplot [semithick, black]
table {%
0.775 1.5625
1.225 1.5625
};
\addplot [semithick, black]
table {%
1.775 0.390625
2.225 0.390625
};
\addplot [semithick, black]
table {%
2.775 3.7112737542943
3.225 3.7112737542943
};
\addplot [semithick, black]
table {%
3.775 0.768610777990738
4.225 0.768610777990738
};
\draw (axis cs:1.5,6.7) node[
  scale=0.8,
  anchor=center,
  text=black,
  rotate=0.0
]{Synthetic data, $\alpha=2$};
\draw (axis cs:3.5,6.7) node[
  scale=0.8,
  anchor=center,
  text=black,
  rotate=0.0
]{German credit dataset};
\end{axis}

\end{tikzpicture}
\caption{Recourse cost by counterfactual fairness outcome.}
\label{fig-boxplot-cost-CF}
\end{subfigure}
\caption{Comparison with counterfactual fairness.}
\label{fig-comparison-CF}
\end{figure}

We observe that the notions of counterfactual fairness and of equality of effort are complementary as each offers different insights into potential unfairness.
Figure \ref{fig-CF-score} shows results of counterfactual fairness ratio (CFR), i.e. the ratio of individuals towards whom our classification algorithm is counterfactually fair as in \cite{ChocklerH2022:aaai}, for some configurations of synthetic data and the German credit dataset. 
Note that, as an inverse to ACR in Figure \ref{fig-ratio-cost-synthetic-alpha}, the CFR is a monotonically decreasing function of $\alpha$, because it is a direct measure of fairness (as opposed to unfairness).
Figure \ref{fig-boxplot-cost-CF} shows the distribution of recourse costs, if we consider the subsets of counterfactually fairly treated individuals (CF) and counterfactually unfairly treated individuals (no CF) for both synthetic data with $\alpha=2$ and for the German credit dataset. The costs are considerably lower under counterfactual unfairness, because counterfactually unfairly treated individuals tend to be closer to the decision boundary as their counterfactual twin would cross the boundary.

\section{Conclusion}\label{sec-conclusion}
We presented a novel approach to assess equality of effort, based on the cost of algorithmic recourse via minimal interventions.
The proposed algorithm has several advantages: the flexibility of the recourse actions (multiple variables can be actionable at the same time) and of the feasibility and plausibility constraints that can be imposed on them; the cost of recourse can be derived directly from the optimization problem; finally relative costs of variables could be directly included in the model as weights in the cost expression, thus capturing a real cost or effort for each intervention. 
We discussed how this notion relates to other causal notions of fairness (equality of effort in~\cite{Huan2020:webconf} and counterfactual fairness of prediction and recourse in~\cite{KusnerLRS2017:anips, ChocklerH2022:aaai}) and \cite{VonKugelgen2022:aaai}, respectively. We applied this notion to synthetic data as well as real data from the German credit dataset, averaging results for different values of hyperparameters and showing equality and inequality of effort at both the individual and the system level.
Our idea can be developed further by including plausibility constraints on covariates and relative costs on the actionable variables in the recourse problem; by enforcing more complex feasibility constraints and taking into account the Recourse Discrepancy in Definition \ref{def-RD}; and by studying the complexity of the algorithm and introducing matching techniques to deal with large datasets. 

\clearpage
\printbibliography

@article{PessacS2020:arxiv,
  title={Algorithmic fairness},
  author={Pessach, Dana and Shmueli, Erez},
  journal={arXiv preprint arXiv:2001.09784},
  year={2020}
}

@article{Makhlouf2020:arxiv,
  title={Survey on causal-based machine learning fairness notions},
  author={Makhlouf, Karima and Zhioua, Sami and Palamidessi, Catuscia},
  journal={arXiv preprint arXiv:2010.09553},
  year={2020}
}

@inproceedings{ChocklerH2022:aaai,
  title={On Testing for Discrimination Using Causal Models},
  author={Chockler, Hana and Halpern, Joseph Y},
  booktitle={Proceedings of the Thirty-Sixth AAAI Conference on Artificial Intelligence},
  year={2022}
}

@article{KusnerLRS2017:anips,
  title={Counterfactual fairness},
  author={Kusner, Matt J and Loftus, Joshua and Russell, Chris and Silva, Ricardo},
  journal={Advances in neural information processing systems},
  volume={30},
  year={2017}
}

@article{Mehrabi2021:acm,
  title={A survey on bias and fairness in machine learning},
  author={Mehrabi, Ninareh and Morstatter, Fred and Saxena, Nripsuta and Lerman, Kristina and Galstyan, Aram},
  journal={ACM Computing Surveys (CSUR)},
  volume={54},
  number={6},
  pages={1--35},
  year={2021},
  publisher={ACM New York, NY, USA}
}

@article{Chouldechova2020:acm,
  title={A snapshot of the frontiers of fairness in machine learning},
  author={Chouldechova, Alexandra and Roth, Aaron},
  journal={Communications of the ACM},
  volume={63},
  number={5},
  pages={82--89},
  year={2020},
  publisher={ACM New York, NY, USA}
}

@book{Pearl2009:cup,
  title={Causality},
  author={Pearl, Judea},
  year={2009},
  publisher={Cambridge university press}
}

@inproceedings{Huan2020:webconf,
  title={Fairness through equality of effort},
  author={Huan, Wen and Wu, Yongkai and Zhang, Lu and Wu, Xintao},
  booktitle={Companion Proceedings of the Web Conference 2020},
  pages={743--751},
  year={2020}
}

@article{Makhlouf2021:elsevier,
  title={Machine learning fairness notions: Bridging the gap with real-world applications},
  author={Makhlouf, Karima and Zhioua, Sami and Palamidessi, Catuscia},
  journal={Information Processing \& Management},
  volume={58},
  number={5},
  pages={102642},
  year={2021},
  publisher={Elsevier}
}

@InProceedings{Heidari2019:arxiv,
  title = 	 {On the Long-term Impact of Algorithmic Decision Policies: Effort Unfairness and Feature Segregation through Social Learning},
  author =       {Heidari, Hoda and Nanda, Vedant and Gummadi, Krishna},
  booktitle = 	 {Proceedings of the 36th International Conference on Machine Learning},
  pages = 	 {2692--2701},
  year = 	 {2019},
  volume = 	 {97},
  series = 	 {Proceedings of Machine Learning Research},
  publisher =    {PMLR},
}

@inproceedings{Karimi2021:acm,
  title={Algorithmic recourse: from counterfactual explanations to interventions},
  author={Karimi, Amir-Hossein and Sch{\"o}lkopf, Bernhard and Valera, Isabel},
  booktitle={Proceedings of the 2021 ACM conference on fairness, accountability, and transparency},
  pages={353--362},
  year={2021}
}

@article{Zhang2016:arxiv,
  title={A causal framework for discovering and removing direct and indirect discrimination},
  author={Zhang, Lu and Wu, Yongkai and Wu, Xintao},
  journal={arXiv preprint arXiv:1611.07509},
  year={2016}
}

@inproceedings{Zhang2016:ijcai,
  title={Situation Testing-Based Discrimination Discovery: A Causal Inference Approach.},
  author={Zhang, Lu and Wu, Yongkai and Wu, Xintao},
  booktitle={Proceedings of the Twenty-Fifth International Joint Conference on Artificial Intelligence},
  volume={16},
  pages={2718--2724},
  year={2016}
}

@misc{Dua2019:uci,
author = "Dua, Dheeru and Graff, Casey",
year = "2017",
title = "{UCI} Machine Learning Repository",
url = "http://archive.ics.uci.edu/ml",
institution = "University of California, Irvine, School of Information and Computer Sciences" }

@inproceedings{Dwork2012:itcs,
  title={Fairness through awareness},
  author={Dwork, Cynthia and Hardt, Moritz and Pitassi, Toniann and Reingold, Omer and Zemel, Richard},
  booktitle={Proceedings of the 3rd innovations in theoretical computer science conference},
  pages={214--226},
  year={2012}
}

@inproceedings{Galhotra2017:jmfse,
  title={Fairness testing: testing software for discrimination},
  author={Galhotra, Sainyam and Brun, Yuriy and Meliou, Alexandra},
  booktitle={Proceedings of the 2017 11th Joint meeting on foundations of software engineering},
  pages={498--510},
  year={2017}
}

@inproceedings{Karimi2020:pmlr,
  title={Model-agnostic counterfactual explanations for consequential decisions},
  author={Karimi, Amir-Hossein and Barthe, Gilles and Balle, Borja and Valera, Isabel},
  booktitle={International Conference on Artificial Intelligence and Statistics},
  pages={895--905},
  year={2020},
  organization={PMLR}
}

@inproceedings{VonKugelgen2022:aaai,
  title={On the fairness of causal algorithmic recourse},
  author={Von K{\"u}gelgen, Julius and Karimi, Amir-Hossein and Bhatt, Umang and Valera, Isabel and Weller, Adrian and Sch{\"o}lkopf, Bernhard},
  booktitle={Proceedings of the AAAI Conference on Artificial Intelligence},
  volume={36},
  pages={9584--9594},
  year={2022}
}

@article{Gupta2019:arxiv,
  title={Equalizing recourse across groups},
  author={Gupta, Vivek and Nokhiz, Pegah and Roy, Chitradeep Dutta and Venkatasubramanian, Suresh},
  journal={arXiv preprint arXiv:1909.03166},
  year={2019}
}

\end{document}